\documentclass[sigconf]{acmart}
\AtBeginDocument{%
  \providecommand\BibTeX{{%
    \normalfont B\kern-0.5em{\scshape i\kern-0.25em b}\kern-0.8em\TeX}}}

\copyrightyear{2023}
\acmYear{2023}
\setcopyright{acmlicensed}\acmConference[KDD '23]{Proceedings of the 29th
ACM SIGKDD Conference on Knowledge Discovery and Data Mining}{August
6--10, 2023}{Long Beach, CA, USA}
\acmBooktitle{Proceedings of the 29th ACM SIGKDD Conference on Knowledge
Discovery and Data Mining (KDD '23), August 6--10, 2023, Long Beach, CA,
USA}
\acmDOI{10.1145/3580305.3599503}
\acmISBN{979-8-4007-0103-0/23/08}

\usepackage{tabto}
\def\theLetterSpace{0.5pt}
\newcommand\spaceout[2][\theLetterSpace]{%
  \def\LocalLetterSpace{#1}\expandafter\spaceouthelpA#2 \relax\relax}
\def\spaceouthelpA#1 #2\relax{%
  \spaceouthelpB#1\relax\relax%
  \ifx\relax#2\else\ \kern\LocalLetterSpace\spaceouthelpA#2\relax\fi
}
\def\spaceouthelpB#1#2\relax{%
  #1%
  \ifx\relax#2\else
    \kern\LocalLetterSpace\spaceouthelpB#2\relax%
  \fi
}
\usepackage{microtype,textcase}
\usepackage{mathtools,xparse}
\usepackage{amsthm, physics}
\usepackage{amsmath}
\usepackage{algorithm,enumitem}
\usepackage{algpseudocode}
\usepackage{algorithmicx}

\usepackage{multirow}
\usepackage{graphicx}
\usepackage{booktabs}
\usepackage{pifont}
\usepackage{xcolor}
\usepackage{blindtext}
\usepackage{cleveref}
\usepackage{balance}

\definecolor{airforceblue}{rgb}{0.36, 0.54, 0.66}
\definecolor{burntumber}{rgb}{0.54, 0.2, 0.14}

\acmConference[KDD'23]{%
   In Proceedings of the 29th ACM SIGKDD Conference on Knowledge Discovery and Data Mining.
  }{%
  August 06-10, 2023}{%
  Long Beach, CA}

\begin{document}

\title{\spaceout[0pt]{Similarity Preserving Adversarial Graph Contrastive Learning}}

\author{Yeonjun In}
\authornote{Both authors contributed equally to this research.}
\affiliation{%
  \institution{KAIST}
  \city{Daejeon}
  \country{Republic of Korea}}
\email{yeonjun.in@kaist.ac.kr}
  
\author{Kanghoon Yoon}
    \authornotemark[1]
\affiliation{%
  \institution{KAIST}
  \city{Daejeon}
  \country{Republic of Korea}}
\email{ykhoon08@kaist.ac.kr}

\author{Chanyoung Park}
\authornote{Corresponding author.}
\affiliation{%
  \institution{KAIST}
  \city{Daejeon}
  \country{Republic of Korea}}
\email{cy.park@kaist.ac.kr}


\renewcommand{\shortauthors}{Yeonjun In, Kanghoon Yoon, and Chanyoung Park.}
\newcommand{\proposed}{\textsf{SP-AGCL}}

\begin{abstract}
\looseness=-1

Recent works demonstrate that GNN models are vulnerable to adversarial attacks, which refer to imperceptible perturbation on the graph structure and node features.
Among various GNN models, graph contrastive learning (GCL) based methods specifically suffer from adversarial attacks due to their inherent design that highly depends on the self-supervision signals derived from the original graph, which however already contains noise when the graph is attacked.
To achieve adversarial robustness against such attacks, existing methods adopt adversarial training (AT) to the GCL framework, which considers the attacked graph as an augmentation under the GCL framework.
However, we find that existing adversarially trained GCL methods achieve robustness \textit{at the expense of not being able to preserve the node feature similarity.} In this paper, we propose a similarity-preserving adversarial graph contrastive learning (\proposed) framework that contrasts the clean graph with two auxiliary views of different properties (i.e., the node similarity-preserving view and the adversarial view). Extensive experiments demonstrate that~\proposed~achieves a competitive performance on several downstream tasks, and shows its effectiveness in various scenarios, e.g., a network with adversarial attacks, noisy labels, and heterophilous neighbors. Our code is available at \href{https://github.com/yeonjun-in/torch-SP-AGCL}{https://github.com/yeonjun-in/torch-SP-AGCL}.
\end{abstract}



\maketitle

\section{Introduction}

\looseness=-1
A graph is a ubiquitous data structure that appears in diverse domains such as chemistry, biology, and social networks. Thanks to their usefulness, a plethora of studies on graph neural networks (GNNs) have been conducted in order to effectively exploit the node and structural information inherent in a graph. However, real-world graphs are usually large-scale, and it is difficult to collect labels due to the expensive cost. Hence, unsupervised graph representation learning methods such as  \cite{deepwalk, node2vec, sage, line} have received steady attention. 
Most recently, the graph contrastive learning (GCL) framework has taken over the mainstream of unsupervised graph representation learning \cite{dgi, mvgrl, grace, gca, lee2022augmentation, lee2022relational}. GCL aims to learn node representations by pulling together semantically similar instances (i.e., positive samples) and pushing apart different instances (i.e., negative samples) in the representation space. In particular, instance discrimination-based approaches \cite{grace, gca}, which treat nodes in differently augmented graphs as self-supervision signals, are dominant among the recent GCL methods.


Although deep learning-based models on graphs have achieved promising results, recent studies have revealed that GNNs are vulnerable to adversarial attacks \cite{pmlr-dai18b-advattack-on-graph, ijacai-19-pgd-topology-attack-defense}. \textit{Adversarial attack on a graph} refers to imperceptible perturbations on the graph structure and node features that rapidly degrade the performance of GNN models. In other words, even with a slight change in the graph structure (e.g., adding/removing a few edges) and node features, GNN models lose their predictive power\footnote{Following recent studies \cite{kdd20-prognn, rsgnn, pagnn}, we mainly focus on the graph structural attack in this work as it is known to be more effective than the node feature attack.}. Existing studies mainly focus on enhancing the robustness of the GNN models, aiming at facilitating robust predictions given an attacked graph. To this end, they introduce novel GNN architectures \cite{NEURIPS2020_GRAND,NEURIPS2020_ReliableAGG, simpgcn, pagnn} or propose to learn a graph structure by purifying the attacked structure \cite{kdd20-prognn, chen2020iterative, rsgnn}. Although the aforementioned approaches have shown effectiveness in training robust GNN models, most of them rely on the label information of nodes and graphs (i.e., supervised setting), and thus they are not applicable when the label information is absent (i.e., unsupervised setting). However, as most real-world graphs are without any label information and the labeling process is costly, developing unsupervised GNN models that are robust to adversarial attacks is important.
Recently, it has been highlighted that unsupervised methods for graphs are also vulnerable to adversarial attacks \cite{pmlr-19a-advattack-on-nodeembedding,clga}, and some approaches have addressed the robustness of unsupervised GNNs by adopting adversarial training (AT) to the GCL framework \cite{grv,ariel}. 
Their main idea is to find the worst-case perturbations on the graph structure and the node features, and use the attacked graph as an adversarial view to train GNN models such that the learned node representations do not vary much despite the adversarial messages propagated from the perturbed edges. In other words, by considering the adversarial view as an augmentation under the GCL framework, they achieve adversarial robustness against adversarial attacks.

{
However, we find out that an adversarially trained GCL model achieves robustness \emph{at the expense of not being able to preserve the similarity among nodes in terms of the node features}, which is an unexpected consequence of applying AT to GCL models~\cite{grv,ariel}. In other words, nodes with similar features do not have similar representations when AT is applied to GCL models. This is mainly due to the fact that applying AT on GCL models force the representations of nodes in the (original) clean graph to be close to those of the attacked graphs in which nodes with dissimilar features are deliberately forced to be connected via adversarial attacks \cite{pmlr-dai18b-advattack-on-graph,kdd20-prognn}. For this reason, the node representations obtained from adversarially trained GCL models fall short of preserving the 
node feature similarity\footnote{\label{note1}This will be empirically shown in Fig. \ref{fig:observation}(b).} despite being robust to graph structural attacks. 
However, as demonstrated by previous studies \cite{kdd20-prognn, simpgcn, rsgnn}, the node feature information is crucial for defending against graph structure attacks. Moreover, preserving the node feature similarity becomes especially useful for most real-world graphs that are incomplete in nature, where a graph network contains noisy node labels and heterophilous neighbors \cite{dai2021nrgnn, simpgcn}.}

\looseness=-1
In this paper, we propose a similarity-preserving adversarial graph contrastive learning (\proposed) framework that preserves the feature similarity information and achieves adversarial robustness. More precisely,~\proposed~contrasts the (original) clean graph with two auxiliary views of different properties (i.e., node similarity-preserving view and adversarial view). The node similarity-preserving view helps preserve the node feature similarity by providing self-supervision signals generated from the raw features of nodes. It is important to note that as the node similarity-preserving view is constructed solely based on the raw features of nodes regardless of the graph structure (i.e., structure-free), the model is particularly robust against graph structural attacks. 
\looseness=-1

Besides, to further exploit the node feature information, we additionally generate an adversarial view by introducing an adversarial feature mask. Specifically, we mask the node features that greatly change the original contrastive loss to generate the adversarial view, which is then used as another view in the GCL framework. The main idea is to learn node representations that are invariant to masked features, which encourages the GCL model to fully exploit the node feature information.
In summary, the main contributions of this paper are three-fold.

\begin{itemize}[leftmargin=0.5cm]
    \item We conduct both theoretical and empirical studies to show that adversarially trained GCL models indeed fail to preserve the node feature similarity. 
    \item We present a novel GCL framework, called \proposed, that achieves adversarial robustness, while preserving the node feature similarity by introducing a similarity-preserving view, and an adversarial view generated from an adversarial feature mask.
    \item\proposed~achieves a competitive performance compared with state-of-the-art baselines on several downstream tasks, and we show its effectiveness in various scenarios, e.g., adversarial attacks, noisy node labels, and heterophilous neighbors.
\end{itemize}


\vspace{-1ex}
\section{Related Works}

\subsection{Graph Contrastive Learning}
Graph Contrastive Learning (GCL) is a well-known representation learning framework that pulls semantically similar instances and pushes semantically different instances. Inspired by Deep InfoMax \cite{dim}, DGI \cite{dgi} presents a contrastive learning method that learns node representations by maximizing the mutual information between the local patch and the global summary of a graph. 
Recently, GRACE \cite{grace} and GCA \cite{gca}, motivated by SimCLR \cite{simclr}, present methods that contrast two differently augmented views with each other, where the different views are generated by various augmentations to edges and node features. GRACE obtains the representations by maximizing the agreements between the representations of the same nodes in the two augmented views, and minimizing the agreements between all other nodes. In this paper, we develop a method that makes GCL models robust against adversarial attacks.


\vspace{-1ex}
\subsection{Adversarial Attacks on Graph}
\looseness=-1
Deep learning models on graphs have been shown to be vulnerable to adversarial attacks. \emph{Nettack} \cite{nettack} is a targeted attack that aims to fool the predictions for specific target nodes. 
\emph{Metattack} \cite{metattack} presents a non-targeted attack method based on meta-learning that deteriorates the overall performance of GNNs. PGD and min-max methods \cite{ijacai-19-pgd-topology-attack-defense} employ loss functions such as negative cross-entropy and CW-type loss to perturb the graph structure using projected gradient descent. However, these attack methods are designed for supervised learning, and thus not applicable under unsupervised settings. 
Most recently, CLGA \cite{clga} introduces an unsupervised graph structural attack method that flips the edges based on the largest gradient of the GCL objectives. In this paper, we focus on developing a GCL framework based on AT, where such adversarially attacked graphs are used as an augmentation to obtain adversarial robustness.

\vspace{-1ex}
\subsection{Adversarial Defense on Graph} 
Along with the growing interest in adversarial attacks on a graph, research on defense methods have also received attention. A strategy for defending the adversarial attack can be categorized into two types. The first line of research focuses on designing novel GNN architectures. Specifically, they achieve robustness by adjusting messages from other nodes based on uncertainty~\cite{kdd-19-rgcn} or node feature similarity information~\cite{simpgcn}.
Another line of research on the adversarial defense focuses on purifying the graph structure by removing noisy edges. They denoise the graph structure with the feature smoothness~\cite{kdd20-prognn} or limited label information~\cite{rsgnn}.
However, the aforementioned approaches are heavily dependent on the supervision signals (i.e., node labels), and thus their performance easily deteriorates with noisy labels. Most importantly, they are not applicable when the label information is absent. 

Training robust GNN models in an unsupervised manner is even more challenging because these models receive (self-) supervision signals from the augmented views of the original graph that may have been already attacked before being augmented, which would not provide helpful supervisory signals. 
Most recently, adversarial GCL models, which apply AT to the GCL framework, have been adopted to unsupervised adversarial defense models. DGI-ADV \cite{grv} 
adopts a training strategy that alternately performs DGI and AT according to the vulnerability of graph representations. ARIEL \cite{ariel} incorporates AT into GRACE, where adversarial augmentation views are generated by adversarial attacks on graph. 
However, these works sacrifice the node feature similarity to obtain robustness against the attack. 
To this end, we propose an adversarial GCL scheme that preserves the node feature similarity.


\begin{figure*}[t]
    
    \centering
    \includegraphics[width=1.75\columnwidth]{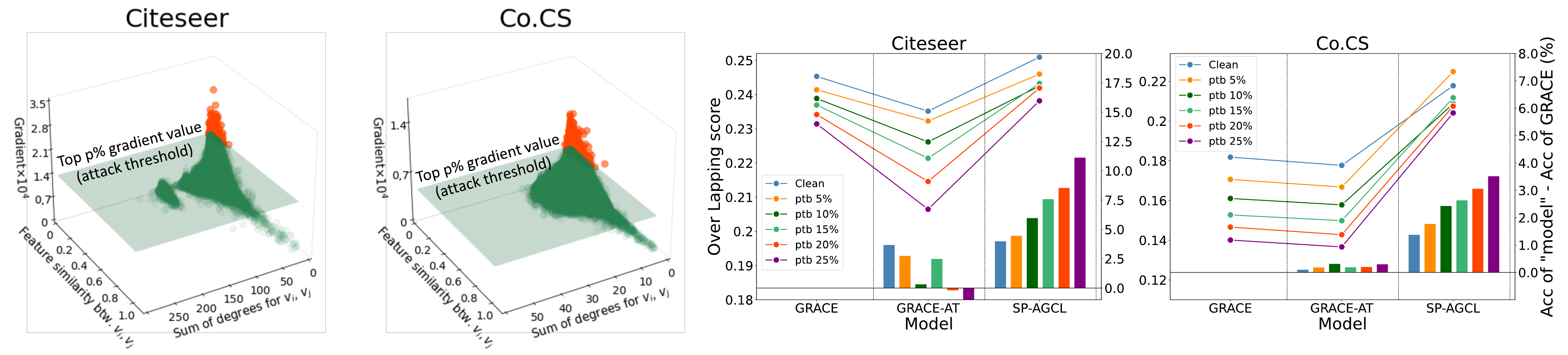}
    \vspace{-2ex}
    \caption{(a) Gradients of the contrastive loss w.r.t $\mathbf{A}$. Each point represents an element in the adjacency matrix $\mathbf{A}$ with the sum of node degrees of two nodes ($x$-axis), and the raw feature similarity ($y$-axis) between two nodes. \textcolor{red}{Red} points denote edges selected for perturbations. (b) $OL$ scores (in solid lines) with $k=10$, and the difference of the node classification (bar graph) accuracy of GRACE-AT, and \proposed~ compared with GRACE on Citeseer and Co.CS datasets.}
    \label{fig:observation}
    \vspace{-2ex}
\end{figure*}

\vspace{-1ex}
\section{Preliminary}
\textbf{Notations.} \@  Let us denote $\mathcal{G}= \langle \mathcal{V},\mathcal{E}, \mathbf{X} \rangle$ as a graph, where $\mathcal{V}=\{v_1,...,v_N\}$ is the set of nodes, $\mathcal{E}\in \mathcal{V}\times \mathcal{V}$ is the set of edges, and $N$ is the number of nodes. We denote $\mathbf{A}\in\mathbb{R}^{N\times N}$ as the adjacency matrix with $\mathbf{A}_{ij}=1$ if $v_i$, $v_j$ are connected, otherwise $\mathbf{A}_{ij}=0$. $\mathbf{X}\in\mathbb{R}^{N\times F}$ and $\mathbf{Z}\in\mathbb{R}^{N\times d}$ are the feature and the representation matrix of nodes, respectively, where $F$ and $d$ are the dimensions of the feature and representation, respectively, and $\mathbf{x}_i$ and $\mathbf{z}_{i}$ denote $i$-th row of $\mathbf{X}$ and $\mathbf{Z}$, respectively. 

\smallskip
\looseness=-1
\noindent \textbf{Graph Contrastive Learning.} \@ 
Since ~\proposed~follows a procedure similar to recent GCL models, we begin by explaining the GCL framework. Particularly, we focus on the instance discrimination models that try to maximize the agreement of representations between different views \cite{grace,gca}. To be specific, given the input $(\mathbf{A},\mathbf{X})$, 
GCL models first generate two different views $(\mathbf{A}^1, \mathbf{X}^1)$ and $(\mathbf{A}^2, \mathbf{X}^2)$ by randomly dropping edges and node features. Then, a GCN \cite{Kipf:2016-gcn} layer $f: (\mathbf{A},\mathbf{X})\rightarrow \mathbf{Z}$ encodes these views into the representations $\mathbf{Z}^1=f(\mathbf{A}^1, \mathbf{X}^1)$ and $\mathbf{Z}^2=f(\mathbf{A}^2, \mathbf{X}^2)$. Finally, the contrastive loss is optimized to enforce the representations of the same nodes in different views to be close, and the representations of different nodes to be distant. Following the common multi-view graph contrastive learning, the pairwise contrastive loss~\cite{grace, gca} is given by: 

\vspace{-3ex}
\begin{equation}
\small
    \mathcal{L}(\mathbf{Z}^1, \mathbf{Z}^2) = \frac{1}{2N}\sum_{i=1}^{N} l(\mathbf{z}_i^1, \mathbf{z}_i^2) + l(\mathbf{z}_i^2, \mathbf{z}_i^1)
    \label{eq:cl_loss}
    \vspace{-3ex}
\end{equation}
\noindent where $l(\mathbf{z}_{i}^{1}, \mathbf{z}_{i}^{2}) = \log \frac{e^{\theta(\mathbf{z}_{i}^{1}, \mathbf{z}_{i}^{2})/\tau}}{e^{\theta(\mathbf{z}_{i}^{1}, \mathbf{z}_{i}^{2})/\tau} + \sum_{k\neq i} e^{\theta(\mathbf{z}_{i}^{1}, \mathbf{z}_{k}^{2})/\tau} + \sum_{k\neq i}e^{\theta(\mathbf{z}_{i}^{1}, \mathbf{z}_{k}^{1})/\tau}}$, and $\theta(z_i^1,  z_i^2)= \frac{g(z_i^1) \cdot g(z_i^2)}{\norm{g(z_i^1)} \norm{g(z_i^2)}}$ is the cosine similarity function with a 2-layer MLP $g$, and $\tau$ is the temperature parameter that scales the distribution. Optimizing the above loss encourages semantically similar nodes to be pulled together and different nodes to be apart in the representation space.

\section{Analysis on Adversarial GCL}
\label{sec:4}

In this section, we perform a theoretical analysis and an empirical study to show that adversarial GCL models indeed fail to preserve the node feature similarity. First, we define adversarial attacks on GCL models and describe how AT is applied to train robust GNN models given an attacked graph (\textbf{Section \ref{subsec:4.1}}). Then, we theoretically analyze how a graph can be effectively attacked, followed by an empirical study to show the characteristics of adversarial attacks on graphs (\textbf{Section \ref{subsec:4.2}}). Lastly, based on the characteristics of adversarial attacks and the AT procedure on GCL, we describe a limitation of existing adversarial GCL models (\textbf{Section \ref{subsec:4.3}}).


\subsection{Applying AT on GCL models}
\label{subsec:4.1}

\noindent \textbf{Adversarial Attack on GCL.} \@ We start with the formulation of the adversarial attack in GCL models. Note that we only consider unsupervised adversarial attacks, where a contrastive loss is employed instead of a supervised loss. Adversarial attacks based on the contrastive loss are conducted as follows:
\begin{equation}
\small
    \delta_{\mathbf{A}}^{*}, \delta_{\mathbf{X}}^{*} = \arg\max_{\delta_{\mathbf{A}}, \delta_{\mathbf{X}}\in \Delta} \mathbb{E}  \left[{ \mathcal{L}( f(\mathbf{A}^1 +
    \delta_{\mathbf{A}}, \mathbf{X}^1 + \delta_{\mathbf{X}}), f(\mathbf{A}^2, \mathbf{X}^2))} \right]
    \label{eq:attack_gcl}
\end{equation}
\noindent where $\Delta=\{(\delta_{\mathbf{A}}, \delta_{\mathbf{X}})| \norm{\delta_{\mathbf{A}}}_{0}\leq \Delta_{\mathbf{A}}, \norm{\delta_{\mathbf{X}}}_{0} \leq \Delta_{\mathbf{X}} \}$ is the possible set of the perturbations, and $\Delta_{\mathbf{A}}$, $\Delta_{\mathbf{X}}$ are perturbation budgets for the edge perturbations $\delta_{\mathbf{A}}$ and the feature perturbations $\delta_{\mathbf{X}}$, respectively. When $\mathbf{A}$ is a discrete matrix and $\mathbf{X}$ consists of binary variables, we use the $l_0$-norm (i.e., count the number of nonzero elements of a vector) for the distance of $\delta_{\mathbf{A}}$ and $\delta_{\mathbf{X}}$. When $\mathbf{X}$ is continuous variables, we use $l_{\infty}$-norm. Hence, the above formulation aims to find the optimal edges and node features to flip (i.e., $\delta_{\mathbf{A}}^*$ and $\delta_{\mathbf{X}}^*$) for the first view that maximally increase the contrastive loss, which in turn makes the representation of the attacked view $f(\mathbf{A}^1+\delta^*_{\mathbf{A}}, \mathbf{X}^1 + \delta^*_{\mathbf{X}})$ to be dissimilar from the representation of the clean view $f(\mathbf{A}^2, \mathbf{X}^2)$. This formulation can be applied to the second view, but here we use the above equation for simplicity.

\smallskip

\noindent \textbf{Training Adversarially Robust GCL.} \@ We explain the adversarial GCL procedure whose goal is to learn robust GCL models based on AT. The main idea of the adversarial GCL is to force the representations of nodes in the clean graph to be close to those of the attacked graphs. Using the attacked graph as an additional augmentation, the adversarial GCL minimizes the training objective 
\begin{equation}
\small
    \min_{\Theta}  { \mathcal{L}(\mathbf{Z}^1,\mathbf{Z}^2) + \lambda_1 \mathcal{L}(\mathbf{Z}^1, \mathbf{Z}^{\text{adv}})}
    \label{eq:at}
\end{equation}
 where $\mathbf{Z}^{\text{adv}} = f(\mathbf{A}^1 + \delta_{\mathbf{A}}^*, \mathbf{X}^1 + \delta_{\mathbf{X}}^*)$ is the representation of the attacked graph with optimal perturbations, $\Theta$ is the set of model parameters, and $\lambda_1$ is a hyperparameter. We hereafter denote the optimally attacked graph $(\mathbf{A}^1 + \delta^*_{\mathbf{A}}, \mathbf{X}^1 + \delta^*_{\mathbf{X}})$ by the \emph{adversarial view}. The first term in the Eqn. (\ref{eq:at}) is the contrastive loss for the two stochastically augmented views, and the second term is the AT loss for learning robust node representations. This adversarial GCL framework regularizes the representation of the original view to be consistent with the representation of the adversarial view.
\looseness=-1
\subsection{\spaceout[-0.08pt]{Characteristic of Adversarial Attacks on GCL}}
\label{subsec:4.2}

Based on the above formulation, we theoretically analyze how a graph can be effectively attacked, which results in degrading the performance of GCL models. Note that adversarial attacks on GNNs under the supervised setting tend to connect nodes with dissimilar features, and it has been shown to greatly change the model prediction results \cite{kdd20-prognn}. Here, we show that similar attack strategies are still effective in degrading the performance of GCL models, which are trained in an unsupervised manner.

\smallskip
\subsubsection{Analyses on effective graph attacks}~\\
\noindent \textbf{1) Theoretical Analysis.} \@
\looseness=-1
Consider a GCL model with a 1-layer GCN encoder without nonlinearity, and assume that there is an arbitrary edge perturbation connecting $v_i$ and $v_k$ where $k \notin \mathcal{N}_{\mathbf{A}}^i$, which is the neighbor set of $v_i$ given the adjacency matrix $\mathbf{A}$. Our goal is to find a single perturbation that greatly increases the contrastive loss 
\footnote{In this analysis, we only consider the structural attack (i.e., $\delta_A$) for simplicity.}. Following Eqn. (\ref{eq:attack_gcl}), this goal can be cast as a problem of finding the optimal perturbation $\delta_{\mathbf{A}}^*$ that makes the representation of the attacked view $\mathbf{Z}^{atk}=f(\mathbf{A}^1 + \delta_{\mathbf{A}}, \mathbf{X})$ dissimilar from those of the second clean view $\mathbf{Z}^2$. Then, the difference in the representations between the clean and the attacked views for node $v_i$ is computed as follows:
\vspace{-2ex}
\begin{align}
\small
& \mathbf{z}_i^2 - {\mathbf{z}_i^{\text{atk}}} = (\mathbf{z}_i^2 - \mathbf{z}_i^1) + (\mathbf{z}_i^1 - {\mathbf{z}_i^{\text{atk}}}) \nonumber \\ 
&= \mathbf{e}_i \!+ \!\!\!\!\!\sum_{j\in\mathcal{N}_{\mathbf{A}^1}^i\cup \{i\}} \frac{{\mathbf{W}\mathbf{x}_j}}{\sqrt{|\mathcal{N}_{\mathbf{A}^1}^i|}\sqrt{|\mathcal{N}_{\mathbf{A}^1}^j|}} - \!\!\!\!\!\sum_{j\in  \mathcal{N}_{\mathbf{A}^1+\delta_{\mathbf{A}}}^i \!\!\!\cup \{i\}} \frac{{\mathbf{W}\mathbf{x}_j}}{\sqrt{|\mathcal{N}_{\mathbf{A}^1+\delta_{\mathbf{A}}}^i|}\sqrt{|\mathcal{N}_{\mathbf{A}^1+\delta_{A}}^j|}} \nonumber \\
&= \mathbf{e}_i + \frac{1}{\underbrace{\sqrt{|\mathcal{N}_{\mathbf{A}^1}^i|+1}}_{\text{Degree term}}} \underbrace{\left(\sum_{j\in\mathcal{N}_{\mathbf{A}^1}^i\cup \{i\}} \frac{\alpha{\mathbf{W}\mathbf{x}_j}}{\sqrt{|\mathcal{N}_{\mathbf{A}^1}^i|}\sqrt{|\mathcal{N}_{\mathbf{A}^1}^j|}} - \frac{\mathbf{W}{\mathbf{x}_k}}{\sqrt{|\mathcal{N}_{\mathbf{A}^1}^k|+1}}\right)}_{\text{Feature difference term}}
\label{eq:diff}
\raisetag{20pt}
\end{align}


\noindent where $\mathbf{e}_i=\mathbf{z}_i^2-\mathbf{z}_i^1$ is the difference between the two clean views, $\alpha$ is a small constant less than $1$, and $\mathbf{W}\in\mathbb{R}^{d\times F}$ is the feature transformation matrix in the GCN encoder. Note that the trained GCL model discriminates two different instances, and $\mathbf{e}_i$ is close to a zero vector as the difference between the two clean views is negligible (i.e., $\mathbf{e}_i\approx \mathbf{0}$). From Eqn. (\ref{eq:diff}), we observe that the distance between the node representations $\mathbf{z_i^{\text{atk}}}$ and $\mathbf{z_i^2}$ is large 1) when the degree of an attacked node $i$ (i.e., $|\mathcal{N}_{\mathbf{A}^1}^i|$) is small, and/or 2) when the feature of the added node through perturbation (i.e., $\mathbf{W}\mathbf{x}_k$) is very different from the aggregation of the neighbor features of $v_i$ (i.e., 1 layer diffusion for $v_i$). This implies that GCL models are vulnerable to adversarial attacks that connect two nodes that are of \emph{low-degree} and exhibit \emph{low feature similarity} with each other. 

\smallskip
\noindent \textbf{2) Empirical Study.} \@
We further conduct an empirical study to verify whether such perturbations are indeed effective in attacking GCL models. We first train a simple GCL model (i.e., GRACE) on Citeseer and Co.CS datasets. Then, since edges (i.e., elements of $\mathbf{A}$) with high gradients greatly change the loss, we visualize the gradient values of the contrastive loss with respect to $\mathbf{A}$ as shown in Fig. \ref{fig:observation}(a). As expected, we observe that the adversarial attacks on GRACE are mainly generated between low-degree and dissimilar nodes, which corroborates our theoretical analysis.

\subsection{Limitation of Existing Adversarial GCL}
\label{subsec:4.3}

As previously demonstrated, adversarial attacks on graphs tend to connect nodes with dissimilar features. Given such an attacked graph, an adversarial GCL model aims to learn robust node representations by reducing the distance between the clean view and the adversarial view (Eqn. (\ref{eq:attack_gcl}),(\ref{eq:at})), where the adversarial view contains more edges that connect nodes with dissimilar features (Eqn. (\ref{eq:diff})). However, although the perturbations in the adversarial view are imperceptible, the neighborhood feature distribution of each node has changed due to the aforementioned characteristic of adversarial attacks on graphs. Hence, we argue that as existing adversarial GCL models force the clean view to be close to the adversarial view while neglecting such changes in the neighborhood feature distributions in the adversarial view, they obtain robustness at the expense of losing the feature information, which is an unexpected consequence of applying AT to GCL models.

\smallskip
\noindent \textbf{Empirical Study.} \@ 
To verify our argument, we conduct empirical studies on several recent adversarial GCL models to investigate whether AT indeed fails to preserve the node feature similarity (Fig.~\ref{fig:observation}(b)). To this end, we measure how much feature similarity in the node features $\mathbf{X}$ is preserved in the node representations $\mathbf{Z}$ learned by adversarial GCL model. Specifically, we first construct $k$-nearest neighbor graphs using $\mathbf{X}$ and $\mathbf{Z}$, and denote them by $\mathbf{A}^{k\text{NN}(\mathbf{X})}$ and $\mathbf{A}^{k\text{NN}(\mathbf{Z})}$, respectively, and compute the overlapping (OL) score between the two graphs as follows \cite{simpgcn}:
\begin{equation}
\small
OL(\mathbf{A}^{k\text{NN}(\mathbf{Z})} , \mathbf{A}^{k\text{NN}(\mathbf{X})})=\frac{|\mathbf{A}^{k\text{NN}(\mathbf{Z})} \cap \mathbf{A}^{k\text{NN}(\mathbf{X})}|}{|\mathbf{A}^{k\text{NN}(\mathbf{X})}|}
\label{eq:ol}
\end{equation}
\noindent where $|\mathbf{A}^{k\text{NN}(\mathbf{X})}|$ is the number of nonzero elements in $\mathbf{A}^{k\text{NN}(\mathbf{X})}$, and $\mathbf{A}^{k\text{NN}(\mathbf{Z})} \cap \mathbf{A}^{k\text{NN}(\mathbf{X})}$ is the intersection of two matrices, i.e., element-wise product of two matrices. Note that a high $OL$ score indicates a high overlap between two matrices $\mathbf{A}^{k\text{NN}(\mathbf{Z})}$ and $\mathbf{A}^{k\text{NN}(\mathbf{X})}$, which implies that nodes with similar representations also have similar features. In other words, if the $OL$ score is high, the node representations contain more information about the node features.

\smallskip
\noindent \textbf{AT fails to Preserve Node Similarity.} \@ 
\looseness=-1
Fig. \ref{fig:observation}(b) shows the $OL$ score of GRACE and GRACE-AT, which is the adversarially trained GCL model built upon GRACE. We observe that the $OL$ scores of GRACE-AT is lower than those of GRACE, which does not employ AT, across all the perturbation ratios, although they outperform GRACE in terms of node classification accuracy. This implies that the learned representations of adversarial GCL model (i.e., GRACE-AT) fail to preserve the feature information while becoming robust to structural attacks, {which means that AT encourages the learned node representations to contain less feature information. However, as demonstrated by previous studies \cite{kdd20-prognn, simpgcn, rsgnn}, the node feature information is crucial for defending against graph structure attacks, and we argue that 
\textit{the robustness against structure attacks of adversarially trained GCL model can be further enhanced fully exploiting the node feature information}.
Hence, in the following section, we propose an adversarial GCL framework that aims to preserve the node feature similarity information while being robust to adversarial attacks. Hence, in the following section, we propose an adversarial GCL framework that aims to preserve the node feature similarity information while being robust to adversarial attacks.}


\section{Proposed Method:~\proposed}
In this section, we propose \textbf{S}imilarirty \textbf{P}reserving \textbf{A}dversarial \textbf{G}raph \textbf{C}ontrastive \textbf{L}earning (\proposed), a framework for the robust unsupervised graph representation learning.~\proposed~achieves adversarial robustness, while preserving the node feature similarity by introducing two auxiliary views for contrastive learning, i,e. the similarity-preserving view and the adversarial view.

\smallskip
\noindent \textbf{Model Overview.} \@
The overall architecture of~\proposed~is described in Fig. \ref{app-fig:architecture} of Appendix~\ref{app-sec:overall_framework}. First, ~\proposed~generates four views that contain different properties: two stochastically augmented views, one similarity preserving view for retaining the feature similarity, and one adversarial view for achieving the adversarial robustness against graph attacks. We then encode the views with GCN layers. Finally, we optimize the cross-view contrastive learning objective based on the encoded views. By optimizing the cross-view objective, the learned node representations obtain the adversarial robustness and enriched feature information.

\subsection{View Generation}
Given a graph $\mathbf{(A,X)}$, we first generate two stochastically augmented views $(\mathbf{A}^1, \mathbf{X}^1)$ and $(\mathbf{A}^2,\mathbf{X}^2)$ both of which are processed by randomly dropping edges and node features as in \cite{grace}. Then, we construct auxiliary views, i.e., the similarity-preserving view and the adversarial view. 
\subsubsection{Similarity-preserving view}
The similarity-preserving view aims to preserve the node similarity information in terms of node features (i.e., $\mathbf{X}$). We first construct a top-$k$ similarity matrix (i.e., ${\mathbf{A}}^{k\text{NN}(\mathbf{X})}$) based on $\mathbf{X}$ using the $k$-nearest neighbor algorithm \cite{knn1, knn2}. More precisely, we select the top-$k$ most similar nodes for each node based on the node feature matrix $\mathbf{X}$. 
The main reason for introducing the similarity-preserving view (i.e., $({\mathbf{A}}^{k\text{NN}(\mathbf{X})}, \mathbf{X})$) is to provide self-supervision signals regarding the node feature information to other views, so that the representations of nodes with similar features are pulled together, which in turn preserves the node feature similarity. This is in contrast to existing adversarial GCL models that obtain robustness against adversarial attacks at the expense of losing the node feature similarity information. 
It is important to note that as the node similarity-preserving view is constructed solely based on the raw features of nodes regardless of the graph structure (i.e., structure-free), the model is particularly robust against structure poisoning attacks . As a result,~\proposed~ is relatively robust even on severly attacked graph, as will be later shown in the experiments.


\subsubsection{Adversarial view}
\label{sec:adv_view}
\looseness=-1
The adversarial view $(\mathbf{A}^{\text{adv}}, \mathbf{X}^{\text{adv}})$ is an augmented view generated by attacking the clean view $(\mathbf{A}^1, \mathbf{X}^1)$ following Eqn. (\ref{eq:attack_gcl}). 
We herein focus on generating an adversarial view that further exploits the node feature information to achieve robustness of an adversarially trained model. In this regard, we present the adversarial feature mask that is applied along with the structural perturbations.
The most common approach for finding the structural and feature perturbations (i.e., $\delta_{\mathbf{A}}^*$ and $\delta_{\mathbf{X}}^*$) is to utilize the gradient-based perturbations that greatly change the contrastive loss. 

\smallskip

\noindent \textbf{1) Structural Perturbations.} \@ For structural perturbations, we follow the similar procedure as \cite{clga} to flip edges by computing the following gradients: $\frac{\partial \mathcal{L}}{\partial \mathbf{A}^1} = \frac{\partial \mathcal{L}}{\partial \mathbf{Z}^1}
    \frac{\partial f(\mathbf{A}^1, \mathbf{X}^1)}{\partial \mathbf{A}^1}$ and $\frac{\partial \mathcal{L}}{\partial \mathbf{A}^2} = \frac{\partial \mathcal{L}}{\partial \mathbf{Z}^2}
    \frac{\partial f(\mathbf{A}^2, \mathbf{X}^2)}{\partial \mathbf{A}^2}$, where $\mathcal{L}$ is the first term in Eqn.~(\ref{eq:at}).
Based on the sum of the gradients (i.e., $\frac{\partial \mathcal{L}}{\partial \mathbf{A}^1}+\frac{\partial \mathcal{L}}{\partial \mathbf{A}^2}=\mathbf{G}_\mathbf{A}\in\mathbb{R}^{N\times N}$), the optimal edges to flip (i.e., $\delta_{\mathbf{A}}^*$) are determined. 
More precisely, for positive elements of $\mathbf{G}_\mathbf{A}$, we take the corresponding edges with large gradients and add them to $\mathbf{A}^1$ to generate $\mathbf{A}^{\text{adv}}$. Moreover, for negative elements of $\mathbf{G}_\mathbf{A}$, we take the corresponding edges with small gradients, and delete them from $\mathbf{A}$ to generate $\mathbf{A}^{\text{adv}}$. Note that the number of edges to add and delete is within the perturbation budget (i.e., $\norm{\delta_{\mathbf{A}}}_0 \leq \Delta_{\mathbf{A}}$).
\smallskip

    

\noindent \textbf{2) Adversarial Feature Mask.} \@ 
For node feature perturbations, a strategy similar to the structural perturbation can be adopted to flip the node features (i.e., change from 0 to 1 and from 1 to 0) as in \cite{grv,ariel}. However, when the feature flipping strategy is applied to node feature perturbations, the co-occurrence/correlation statistics of nodes are significantly altered~\cite{nettack}. Such a behavior may have an adverse effect on the AT by making the clean view and the adversarial view too distant from each other.
Hence, to perturb node features while retaining the co-occurrence/correlation statistics of node features, we propose to mask (i.e., only change from 1 to 0) features that greatly increase the contrastive loss. 
Specifically, considering that we are interested in changing the 1s in the feature matrix $\textbf{X}$ to 0s, we mask the node features with small gradients in the negative direction, since doing so will greatly increase the loss. 
More formally, to obtain the adversarial mask, we compute the following gradients: $\frac{\partial \mathcal{L}}{\partial \mathbf{X}^1} = \frac{\partial \mathcal{L}}{\partial \mathbf{Z}^1}
    \frac{\partial f(\mathbf{A}^1, \mathbf{X}^1)}{\partial \mathbf{X}^1}$ and $\frac{\partial \mathcal{L}}{\partial \mathbf{X}^2} = \frac{\partial \mathcal{L}}{\partial \mathbf{Z}^2}
    \frac{\partial f(\mathbf{A}^2, \mathbf{X}^2)}{\partial \mathbf{X}^2}$, where $\mathcal{L}$ is the first term in Eqn.~(\ref{eq:at}). Based on the sum of the gradients with respect to the node features (i.e., $\frac{\partial \mathcal{L}}{\partial \mathbf{X}^1}+\frac{\partial \mathcal{L}}{\partial \mathbf{X}^2}=\mathbf{G}_\mathbf{X}\in\mathbb{R}^{N\times F}$), the adversarial feature mask $\mathbf{M}$ is obtained. 
More precisely, for negative $\mathbf{G}_\mathbf{X}$, we take the node features with small gradients to generate the mask $\textbf{M}$. That is, $\mathbf{M}_{ij}=0$ if the $j$-th feature of node $i$ has a small gradient, otherwise $\mathbf{M}_{ij}=1$, where the number of zeros in the mask $\mathbf{M}$ is within the perturbation budget (i.e., $\norm{\mathbf{M}}_0 \leq \Delta_{\mathbf{X}}$). Then, we apply $\mathbf{M}$ to obtain the node features of the adversarial view (i.e., $\mathbf{X}^{\text{adv}}$) as follows:
$\mathbf{X}^{\text{adv}} = \mathbf{M}\odot \mathbf{X}^1$, where $\odot$ is hadamard product for matrices. 
By masking the node features that play an important role in the contrastive learning, and using it as another view in the GCL framework, we expect to learn node representations that are invariant to masked features, which encourages the GCL model to fully exploit the node feature information.

\smallskip
\noindent\textbf{3) Adversarial View Generation.}
Combining the result of structural perturbations (i.e., $\mathbf{A}^{\text{adv}}$) and the result of adversarial feature masking (i.e., $\mathbf{X}^{\text{adv}}$), the adversarial view can be obtained (i.e., $(\mathbf{A}^{\text{adv}}, \mathbf{X}^{\text{adv}})$), and this view
contains both structural and feature perturbations, 
which makes the adversarial view more helpful for achieving adversarial robustness.

\begin{table*}[t!]
\caption{Node classification accuracy under non-targeted attack (\emph{metattack}). (OOM: Out of Memory on 24GB RTX3090).}
\vspace{-2ex}
\renewcommand{\arraystretch}{1}
\begin{center}
{\scriptsize
\scalebox{0.85}{%
\begin{tabular}{c|l|c|ccccc|ccccc}
    \toprule
    \multirow{1}{*}{} & \multirow{1}{*}{Methods}& {} & \multicolumn{5}{c}{Poisoning (Acc.)} & \multicolumn{5}{c}{Evasive (Acc.)} \\
    \hline
    {Datasets}&{Ptb rate}&{Clean}&{5\%}& 10\% & 15\% & 20\% & 25\% & 5\% & 10\% & 15\% & 20\% & 25\% \\
    \midrule
    \multirow{5}{*}{Cora} & GRACE-MLP & 63.2±1.7 & 63.2±1.7 & 63.2±1.7 & 63.2±1.7 & 63.2±1.7 & 63.2±1.7 & 63.2±1.7 & 63.2±1.7 & 63.2±1.7 & 63.2±1.7 & 63.2±1.7 \\ 
    & GRACE   
    & 82.1±1.0            & 78.4±1.5          & 75.5±1.1          & 66.1±1.6          & 55.2±1.8          & 51.3±2.0            & 78.9±0.9          & 75.7±0.9          & 67.6±1.3          & 56.5±2.3          & 51.5±1.8   \\
    & GCA     
    & 81.5±0.9          & 79.8±0.8          & 75.8±0.6          & 68.4±1.6          & 53.4±1.7          & 49.5±1.3          & 79.7±1.0            & 76.0±1.1            & 68.0±1.1            & 54.7±1.2          & 49.8±1.3       \\
    & BGRL   
    & 82.7±1.0            & 78.2±2.1          & 74.3±1.8          & 66.2±1.9          & 53.8±1.7          & 50.2±2.3          & 79.2±1.6          & 75.2±1.5          & 67.2±2.0            & 55.2±1.7          & 51.2±1.7    \\
    & DGI-ADV 
    & 83.7±0.7          & 79.4±0.9          & 73.3±0.6          & 63.5±0.6          & 52.2±0.7          & 48.1±0.7          & 79.4±0.9          & 73.7±0.8          & 62.9±0.9          & 53.0±1.0              & 49.2±1.2          \\
    & ARIEL   
    & 80.9±0.5          & 79.2±0.4          & 77.7±0.6          & 69.8±0.7          & 57.7±0.7          & 52.8±1.0            & 79.1±0.3          & 77.8±0.6          & 70.3±0.9          & 58.0±1.0              & 53.2±1.2          \\
    \hline
    
    & \proposed 
    & \textbf{83.9±0.7} & \textbf{82.2±0.8} & \textbf{79.0±0.6} & \textbf{73.25±0.5} & \textbf{66.2±2.3} & \textbf{65.0±1.5} & \textbf{82.0±0.6}   & \textbf{78.7±1.2} & \textbf{73.5±2.8} & \textbf{61.5±5.0}   & \textbf{57.1±5.5} \\
    
    \midrule
    \multirow{5}{*}{Citeseer} 
    & GRACE-MLP & 68.0±1.2 & 68.0±1.2 & 68.0±1.2 & 68.0±1.2 & 68.0±1.2 & 68.0±1.2 & 68.0±1.2 & 68.0±1.2 & 68.0±1.2 & 68.0±1.2 & 68.0±1.2    \\
    & GRACE 
    & 74.9{±0.6}          & 74.1{±0.6}          & 72.5{±0.9}          & 71.2{±1.3}          & 59.2{±1.4}          & 61.2{±1.5}          & 74.0{±0.7}            & 72.4{±1.0}            & 70.4{±1.3}          & 59.1{±1.9}          & 62.3{±1.5}          \\
    & GCA
    & 74.2{±0.7}          & 73.5{±0.9}          & 73.0{±0.6}            & 71.5{±0.9}          & 60.2{±1.7}          & 60.1{±1.6}          & 73.8{±0.7}          & 73.4{±0.5}          & 72.0{±0.9}            & 59.5{±1.8}          & 61.5{±1.7}          \\
    & BGRL    
    & 73.4{±1.0}            & 72.1{±1.1}          & 69.1{±1.0}            & 67.5{±1.4}          & 57.7{±1.3}          & 58.2{±2.8}          & 72.5{±1.2}          & 69.7{±1.3}          & 68.1{±1.6}          & 58.5{±1.6}          & 60.3{±2.0}            \\
    & DGI-ADV 
    & 76.6{±0.3}          & 74.8{±0.3}          & 71.0{±0.5}            & 70.1{±0.3}          & 57.9{±0.8}          & 60.6{±1.2}          & 74.8{±0.3}          & 71.3{±0.5}          & 69.7{±0.5}          & 56.1{±0.6}          & 57.4{±1.5}          \\
    & ARIEL   
    & \textbf{76.7{±0.5}} & {75.2{±0.4}} & 72.8{±0.5}          & 70.2{±0.5}          & 60.1{±1.1}          & 62.7{±0.5}          & \textbf{75.3{±0.4}} & 73.3{±0.5}          & 70.8{±0.4}          & 59.8{±0.8}          & 63.6{±1.0}            \\
    \hline
    & \proposed  
    & 75.9{±0.4}          & \textbf{75.3±0.5} & \textbf{73.5±0.6} & \textbf{72.1±1.1} & \textbf{66.0±1.5} & \textbf{69.6±0.9} & 75.0{±1.1}            & \textbf{73.5{±1.0}}   & \textbf{72.4{±1.1}} & \textbf{60.6{±1.1}} & \textbf{65.6{±0.9}} \\
    
    \midrule
    \multirow{5}{*}{Pubmed} 
    & GRACE-MLP 
    &  82.4±0.2 & 82.4±0.2 & 82.4±0.2 & 82.4±0.2 & 82.4±0.2 & 82.4±0.2 & 82.4±0.2 & 82.4±0.2 & 82.4±0.2 & 82.4±0.2 & 82.4±0.2      \\ 
    & GRACE 
    & 85.9{±0.1}          & 81.3{±0.2}          & 78.2{±0.4}          & 76.1{±1.3}          & 73.9{±1.7}          & 71.3{±2.6}          & 80.7{±0.1}          & 76.8{±0.2}          & 73.5{±0.1}          & 71.4{±0.2}          & 69.0{±0.3}            \\
    & GCA     
    & \textbf{86.5±{0.2}} & 81.2{±0.5}          & 78.1{±0.5}          & 75.9{±1.2}          & 74.2{±0.4}          & 72.0{±1.8}            & 80.7{±0.2}          & 76.7{±0.3}          & 73.2{±0.3}          & 70.9{±0.2}          & 68.6{±0.3}          \\
    & BGRL    
    & 85.1{±0.2}          & 81.3{±0.3}          & 79.0{±0.4}            & 76.6{±0.9}          & 74.8{±0.9}          & 73.0{±0.5}            & 80.6{±0.4}          & 77.5{±0.4}          & 74.5{±0.6}          & 72.4{±0.7}          & 70.3{±0.6}          \\
    & DGI-ADV 
    & {OOM}               & {OOM}               & {OOM}               & {OOM}               & {OOM}               & {OOM}               & {OOM}               & {OOM}               & {OOM}               & {OOM}               & {OOM}               \\
    & ARIEL   
    & 81.2{±0.4}          & 77.8{±0.3}          & 75.8{±0.4}          & 74.0{±0.5}            & 72.3{±0.5}          & 70.7{±0.3}          & 77.8{±0.5}          & 75.9{±0.5}          & 74.1{±0.6}          & 72.3{±0.6}          & 70.8{±0.5}\\
    \hline              
    & \proposed  
    & 85.5{±0.3}          & \textbf{81.9{±0.2}} & \textbf{80.2±0.1} & \textbf{77.9±0.4} & \textbf{76.5±0.1} & \textbf{73.3±0.3} & \textbf{81.9{±0.2}} & \textbf{79.6{±0.3}} & \textbf{77.1{±0.4}} & \textbf{75.1{±0.5}} & \textbf{72.8{±0.5}} \\
    
    \midrule
    \multirow{5}{*}{Am.Photo} 
    & GRACE-MLP
    & 87.2±0.8 & 87.2±0.8 & 87.2±0.8 & 87.2±0.8 & 87.2±0.8 & 87.2±0.8 & 87.2±0.8 & 87.2±0.8 & 87.2±0.8 & 87.2±0.8 & 87.2±0.8 \\
    & GRACE
    & 92.0{±0.4}          & 89.5{±0.5}          & 88.3{±1.1}          & 87.6{±0.9}          & 87.5{±1.2}          & 87.1{±1.2}          & 88.6{±0.4}          & 87.5{±0.8}          & 87.3{±0.8}          & 86.6{±1.0}          & 85.6{±1.1}          \\
    & GCA     
    & 92.2{±0.4}          & 89.4{±0.6}          & 88.3{±0.8}          & 87.8{±0.7}          & 87.6{±1.0}          & 87.5{±0.7}          & 88.7{±0.6}          & 88.0{±0.7}          & 87.8{±1.4}          & 87.3{±0.9}          & 86.4{±1.2}          \\
    & BGRL    
    & 92.1{±0.4}          & 89.2{±0.6}          & 88.7{±0.5}          & 88.8{±0.5}          & 89.0{±0.7}          & 89.2{±0.4}          & 89.4{±0.5}          & 88.3{±0.6}          & 88.2{±0.6}          & 87.6{±0.5}          & 87.3{±0.6}          \\
    & DGI-ADV 
    & 91.6{±0.5}          & 83.5{±0.5}          & 80.7{±0.6}          & 79.3{±0.6}          & 78.1{±0.6}          & 77.3{±0.6}          & 83.6{±0.5}          & 80.8{±0.5}          & 79.5{±0.5}          & 78.0{±0.6}          & 77.4{±0.4}          \\
    & ARIEL   
    & 92.5{±0.2}          & 90.1{±0.4}          & 89.9{±0.5}          & 89.9±0.5          & 89.9{±0.5}          & \textbf{89.8{±0.6}} & 89.7{±0.4}          & 89.1{±0.3}          & 88.6{±0.2}          & \textbf{88.6{±0.4}} & \textbf{88.4{±0.3}} \\
    \hline              
    & \proposed 
    & \textbf{93.3{±0.3}} & \textbf{91.4{±0.5}} & \textbf{90.6{±0.6}} & \textbf{90.5{±0.8}} & \textbf{90.2{±0.9}} & \textbf{89.8{±1.0}} & \textbf{90.3{±0.4}} & \textbf{89.3{±0.3}} & \textbf{88.7{±0.5}} & 88.2{±0.7} & 87.6{±0.5}\\
    
    \midrule
    \multirow{5}{*}{Am.Comp} 
    & GRACE-MLP
    & 82.7±0.4 & 82.7±0.4 & 82.7±0.4 & 82.7±0.4 & 82.7±0.4 & 82.7±0.4 & 82.7±0.4 & 82.7±0.4 & 82.7±0.4 & 82.7±0.4 & 82.7±0.4          \\
    & GRACE 
    & 86.4{±0.5}          & 83.7{±0.3}          & 82.5{±0.6}          & 81.3{±0.5}          & 80.3{±0.7}          & 78.8{±1.0}          & 84.0{±0.3}          & 83.6{±0.4}          & 82.8{±0.3}          & 82.0{±0.9}          & 81.7{±0.6}          \\
    & GCA     
    & 86.6{±0.4}          & 84.6{±0.4}          & 83.4{±0.3}          & 82.3{±0.4}          & 81.4{±0.4}          & 80.1{±0.5}          & 84.5{±0.3}          & 83.8{±0.4}          & 82.8{±0.2}          & 82.3{±0.8}          & 82.0{±0.4}          \\
    & BGRL    
    & 88.0{±0.4}          & 85.2{±0.6}          & 84.2{±0.6}          & 83.7{±0.6}          & 83.3{±0.7}          & 83.4{±0.6}          & 85.7{±0.7}          & 85.0{±0.6}          & 84.1{±0.6}          & 83.8{±0.7}          & 83.4{±0.5}          \\
    & DGI-ADV
    & {OOM}               & {OOM}               & {OOM}               & {OOM}               & {OOM}               & {OOM}               & {OOM}               & {OOM}               & {OOM}               & {OOM}               & {OOM}               \\
    & ARIEL   
    & 87.4{±0.4}          & 85.4{±0.5}          & 84.5{±0.4}          & 83.8{±0.4}          & 83.7{±0.5}          & 83.6{±0.5}          & 85.7{±0.3}          & 84.6{±0.4}          & 83.9{±0.5}          & 83.8{±0.4}          & 83.6{±0.5}          \\
    \hline              
    & \proposed  
    & \textbf{89.1{±0.4}} & \textbf{86.9{±0.3}} & \textbf{85.6{±0.5}} & \textbf{85.1{±0.4}} & \textbf{85.0{±0.5}} & \textbf{84.8{±0.7}} & \textbf{87.2{±0.3}} & \textbf{85.9{±0.4}} & \textbf{85.1{±0.5}} & \textbf{84.4{±0.4}} & \textbf{84.1{±0.6}} \\
    
    \midrule
    \multirow{5}{*}{Co.CS} 
    & GRACE-MLP 
    & 92.1±0.2 & 92.1±0.2 & 92.1±0.2 & 92.1±0.2 & 92.1±0.2 & 92.1±0.2 & 92.1±0.2 & 92.1±0.2 & 92.1±0.2 & 92.1±0.2 & 92.1±0.2  \\
    & GRACE 
    & 92.3{±0.2}          & 91.2{±0.1}          & 90.6{±0.2}          & 90.0{±0.2}          & 89.4{±0.1}          & 88.9{±0.2}          & 91.2{±0.3}          & 90.4{±0.3}          & 89.7{±0.4}          & 89.3{±0.4}          & 88.7{±0.4}          \\
    & GCA     
    & 92.5{±0.1}          & 91.4{±0.2}          & 90.7{±0.2}          & 90.2{±0.2}          & 89.7{±0.2}          & 89.3{±0.1}          & 91.4{±0.2}          & 90.8{±0.3}          & 90.0{±0.3}          & 89.6{±0.3}          & 89.0{±0.3}          \\
    & BGRL    
    & 92.4{±0.2}          & 91.3{±0.1}          & 90.5{±0.2}          & 89.9{±0.2}          & 89.3{±0.2}          & 88.7{±0.2}          & 91.3{±0.2}          & 90.5{±0.2}          & 89.8{±0.3}          & 89.4{±0.3}          & 88.8{±0.2}          \\
    & DGI-ADV
    & {OOM}               & {OOM}               & {OOM}               & {OOM}               & {OOM}               & {OOM}               & {OOM}               & {OOM}               & {OOM}               & {OOM}               & {OOM}               \\
    & ARIEL   
    & 92.3{±0.2}          & 91.0{±0.1}          & 90.2{±0.2}          & 89.6{±0.3}          & 88.8{±0.2}          & 88.1{±0.2}          & 90.8{±0.1}          & 90.1{±0.3}          & 89.2{±0.2}          & 88.7{±0.2}          & 87.9{±0.1}          \\
    \hline              
    & \proposed 
    & \textbf{93.7±0.2} & \textbf{92.9±0.2} & \textbf{92.8±0.2} & \textbf{92.5±0.2} & \textbf{92.4±0.1} & \textbf{92.3±0.2} & \textbf{92.7±0.1} & \textbf{91.9±0.2} & \textbf{91.2±0.2} & \textbf{90.6±0.2} & \textbf{89.9±0.2} \\
    
    \midrule
    \multirow{5}{*}{Co.Physics} 
    & GRACE-MLP  
    & {OOM}               & {OOM}               & {OOM}               & {OOM}               & {OOM}               & {OOM}               & {OOM}               & {OOM}               & {OOM}               & {OOM}               & {OOM}               \\
    & GRACE  
    & {OOM}               & {OOM}               & {OOM}               & {OOM}               & {OOM}               & {OOM}               & {OOM}               & {OOM}               & {OOM}               & {OOM}               & {OOM}               \\
    & GCA  
    & {OOM}               & {OOM}               & {OOM}               & {OOM}               & {OOM}               & {OOM}               & {OOM}               & {OOM}               & {OOM}               & {OOM}               & {OOM}               \\
    & BGRL    
    & 95.2{±0.1}          & 94.1{±0.2}          & 93.2{±0.2}          & 92.5±0.1          & 91.6{±0.2}          & 91.0{±0.1}          & 94.2{±0.1}          & 93.2{±0.2}          & 92.5{±0.1}          & 91.6{±0.1}          & 91.0{±0.2}          \\
    & DGI-ADV 
    & {OOM}               & {OOM}               & {OOM}               & {OOM}               & {OOM}               & {OOM}               & {OOM}               & {OOM}               & {OOM}               & {OOM}               & {OOM}               \\
    & ARIEL   
    & 95.1{±0.1}          & 93.2{±0.2}          & 92.4{±0.2}          & 91.6{±0.2}          & 90.7{±0.3}          & 90.2{±0.2}          & 93.9{±0.1}          & 93.3{±0.1}          & 92.6{±0.2}          & 91.8{±0.2}          & 91.4{±0.2}          \\
    \hline              
    & \proposed  
    & \textbf{95.8{±0.1}} & \textbf{94.9±{0.2}} & \textbf{94.4{±0.1}} & \textbf{93.6±{0.1}} & \textbf{93.0{±0.1}} & \textbf{92.5±0.1} & \textbf{95.0{±0.1}} & \textbf{94.2{±0.1}} & \textbf{93.3±0.2} & \textbf{92.4{±0.1}} & \textbf{91.7{±0.1}}\\
    \bottomrule
\end{tabular}}}
\end{center}
\vspace{-0ex}
\label{tab:main_table}
\end{table*}

\begin{figure*}[t]
\vspace{-1ex}
  \includegraphics[width=.6\linewidth]{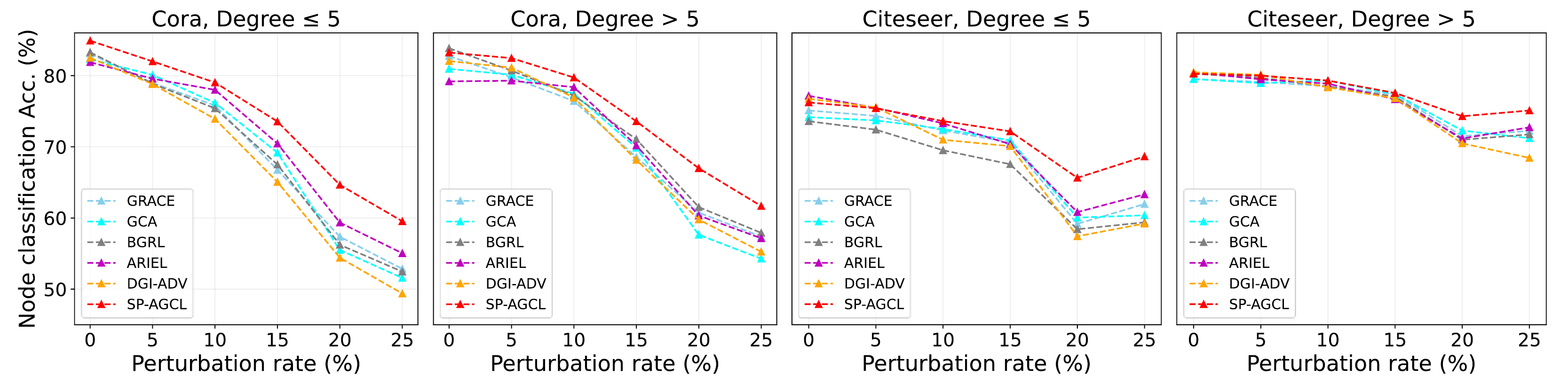}
  \vspace{-1ex}
  \caption{Node classification on low-/high-degree nodes under \emph{metattack}. Cora and Citseer datasets are used.}
  \label{fig:low_degree}
  \vspace{-2ex}
\end{figure*}


\smallskip
\looseness=-1
\noindent \textbf{Applicability to large networks.} \@ 
Since computing the gradient with respect to both $\mathbf{A}$ and $\mathbf{X}$ requires expensive cost, adversarial GCL models are generally not scalable to large graphs. Hence, we generate all the views, i.e., $(\mathbf{A}^1,\mathbf{X}^1),(\mathbf{A}^2,\mathbf{X}^2),({\mathbf{A}}^{k\text{NN}(\mathbf{X})}, \mathbf{X})$, and $(\mathbf{A}^{\text{adv}}, \mathbf{X}^{\text{adv}})$, for only a subset of the original graph by randomly sampling a subgraph $\bar{\mathcal{G}}$ with $(\bar{\mathbf{A}},\bar{\mathbf{X}})$ from the original graph $\mathcal{G}$ \cite{ariel, mvgrl}, and then apply the same view generation procedure described above. 

\subsection{Cross-view Training for Robust GCL}
Based on the representations obtained from different views, we train~\proposed~by using the cross-view contrastive objective:
\begin{equation}
\small
    \min_{\Theta} \mathcal{L}(\mathbf{Z}^1, \mathbf{Z}^2) + \lambda_1 \mathcal{L}(\mathbf{Z}^1, \mathbf{Z}^{\text{adv}}) + \lambda_2 \mathcal{L}(\mathbf{Z}^1, \mathbf{Z}^{k\text{NN}(\mathbf{X})})
\label{final_loss}
\end{equation}
\noindent where $\lambda_1$ and $\lambda_2$ are hyperparameters. The first term trains a GCL model, the second term contrasts a clean view with an adversarial view, and the last term contrasts the clean view with the similarity preserving view. Training the above objective enhances the robustness and preserves the feature similarity of the node representation. We empirically discovered that replacing $\mathbf{Z}^1$ in $\mathcal{L}(\mathbf{Z}^1, \mathbf{Z}^{\text{adv}})$ and $\mathcal{L}(\mathbf{Z}^1, \mathbf{Z}^{k\text{NN}(\mathbf{X})})$ with $\mathbf{Z}^2$ does not make a significant difference to the model performance.

\section{Experiment}

\begin{table*}[t]
\caption{Node classification accuracy under targeted attack (\emph{nettack}). }
\vspace{-3ex}
\renewcommand{\arraystretch}{1}
\begin{center}
{\scriptsize
\scalebox{0.85}{
\begin{tabular}{c|l|c|ccccc|ccccc}
    \toprule
    \multirow{1}{*}{} & \multirow{1}{*}{Methods} & {} & \multicolumn{5}{c}{Poisoning (Acc.)} & \multicolumn{5}{c}{Evasive (Acc.)}  \\
    \hline
    {Datasets}&{\# Ptb}&{Clean}&{1}& 2 & 3 & 4 & 5 &{1}& 2 & 3 & 4 & 5\\
    \midrule
    \multirow{5}{*}{Cora} & GRACE   & 82.2±2.2          & 76.9±1.5          & 70.2±2.4          & 65.9±3.0          & 64.6±1.5          & 58.9±2.1          & 77.7±2.7          & 71.1±2.5          & 67.1±2.5          & 65.1±2.3          & 60.2±3.3          \\
    & GCA     & 81.3±1.7          & 77.7±2.1          & 71.6±2.2          & 67.2±2.3          & 63.9±2.2          & 59.2±2.0          & 79.2±1.2          & 73.0±0.8          & 69.2±1.2          & 67.1±1.5          & 62.3±3.0          \\
    & BGRL    & \textbf{83.0±2.3} & 78.6±1.9          & 73.0±3.9          & 69.3±2.9          & 63.7±5.0          & 60.5±3.1          & 79.2±2.8          & 74.5±2.1          & 70.7±3.2          & 66.9±2.8          & 64.1±3.0          \\
    & DGI-ADV & 81.7±0.7          & 78.0±2.3          & 71.1±2.1          & 69.9±1.1          & 65.7±1.8          & 60.7±1.9          & 78.1±1.6          & 73.0±2.1          & 70.6±1.3          & 66.5±1.2          & 63.4±1.5          \\
    & ARIEL   & 76.0±1.7          & 71.9±2.2          & 64.9±1.3          & 63.5±1.6          & 63.0±1.6          & 53.7±1.7          & 71.7±2.0          & 65.4±1.2          & 63.5±1.5          & 64.1±1.3          & 54.6±1.6          \\
    \hline
    
    & \proposed 
    & 82.5±2.0          & \textbf{79.5±1.8} & \textbf{75.3±2.5} & \textbf{73.4±1.7} & \textbf{67.4±2.3} & \textbf{63.6±2.4} & \textbf{80.2±2.4} & \textbf{78.7±3.1} & \textbf{77.1±3.2} & \textbf{73.5±3.4} & \textbf{72.8±3.7} \\
    
    \midrule
    \multirow{5}{*}{Citeseer} 
    & GRACE   & 82.4±0.5          & 81.8±1.1          & 77.6±4.2          & 68.3±4.4          & 64.3±3.0          & 59.1±2.7          & 82.2±0.6          & 81.1±1.3          & 78.1±3.2          & 72.4±4.8          & 66.4±3.9          \\
    & GCA     & \textbf{82.5±0.0} & 82.4±0.5          & 78.3±2.9          & 69.4±5.9          & 65.9±2.0          & 58.3±4.0          & \textbf{82.5±0.0} & 81.1±1.5          & 79.2±2.5          & 77.0±2.6          & 71.3±4.4          \\
    & BGRL    & \textbf{82.5±0.7} & 81.4±1.2          & 79.7±4.6          & 75.1±7.3          & 72.7±7.6          & 67.3±8.5          & 81.6±1.1          & 80.0±3.3          & 78.9±4.0          & 76.7±5.6          & 73.3±6.5          \\
    & DGI-ADV & \textbf{82.5±0.0} & 81.4±0.7          & 80.2±1.5          & 74.3±3.7          & 68.6±1.2          & 65.6±1.2          & 82.4±0.5          & 81.3±1.0          & 79.7±0.6          & 78.7±1.3          & 76.5±1.6          \\
    & ARIEL   & \textbf{82.5±0.0} & 81.1±0.9          & 80.6±0.6          & 74.3±3.9          & 66.2±1.6          & 63.2±1.0          & 81.9±0.8          & 81.3±0.6          & 81.0±0.0          & \textbf{80.2±0.8} & \textbf{78.6±1.5} \\
    \hline
    & \proposed  
    & \textbf{82.5±0.0} & \textbf{82.5±0.0} & \textbf{81.6±1.1} & \textbf{80.0±3.0} & \textbf{75.4±6.1} & \textbf{72.7±5.3} & 82.4±0.5          & \textbf{82.1±1.0} & \textbf{81.6±1.6} & \textbf{80.2±4.2} & 78.1±4.9          \\
    
    \midrule
    \multirow{5}{*}{Pubmed} 
    & GRACE   & 87.9±0.6          & 86.6±0.5          & 84.3±0.8          & 81.7±0.9          & 77.9±1.5          & 73.0±1.3          & 86.3±0.4          & 84.4±0.7          & 81.8±1.0          & 78.3±1.2          & 74.5±1.3          \\
    & GCA     & \textbf{88.0±0.5} & \textbf{87.1±0.6} & \textbf{84.7±0.7} & 82.2±1.5          & 77.7±1.2          & 73.6±1.7          & \textbf{87,0±0.5} & \textbf{85.2±0.7} & 82.7±1.1          & 79.2±0.7          & 75.5±1.2          \\
    & BGRL    & 87.4±0.8          & 85.8±0.8          & 83.2±1.0          & 79.3±1.0          & 75.4±1.2          & 70.2±1.2          & 85.8±1.1          & 83.7±1.0          & 80.3±1.2          & 76.1±1.0          & 72.3±1.1          \\
    & DGI-ADV & OOM               & OOM               & OOM               & OOM               & OOM               & OOM               & OOM               & OOM               & OOM               & OOM               & OOM               \\
    & ARIEL   & 83.9±0.8          & 82.0±1.0          & 79.4±1.5          & 75.2±1.4          & 72.0±1.3          & 67.5±2.6          & 81.9±0.7          & 78.8±1.6          & 76.1±1.2          & 72.0±1.2          & 68.2±1.6          \\
    \hline              
    & \proposed  
    & 87.4±0.8          & 85.9±0.7          & 84.3±1.0          & \textbf{82.7±1.1} & \textbf{80.1±1.6} & \textbf{77.7±2.2} & 86.2±0.5          & 84.5±0.7          & \textbf{82.9±0.6} & \textbf{81.0±1.1} & \textbf{78.1±1.6} \\
    \bottomrule
\end{tabular}}}
\end{center}
\vspace{-1ex}
\label{tab:main_table2}
\end{table*}

\begin{table*}[t!]
\caption{Node classification accuracy under random perturbations.}
\vspace{-3ex}
\renewcommand{\arraystretch}{1}
\begin{center}
{\scriptsize
\scalebox{0.85}{
\begin{tabular}{c|l|c|ccccc|ccccc}
    \toprule
    \multirow{1}{*}{} & \multirow{1}{*}{Methods} & {} & \multicolumn{5}{c}{Poisoning (Acc.)} & \multicolumn{5}{c}{Evasive (Acc.)}  \\
    \hline
    {Datasets}&{Ptb rate}&{Clean}&{20\%}& 40\% & 60\% & 80\% & 100\%&{20\%}& 40\% & 60\% & 80\% & 100\% \\
    \midrule
    \multirow{5}{*}{Cora} & GRACE   & 82.1±1.0          & 77.5±1.2          & 74.2±0.9          & 70.3±1.2          & 66.9±1.1          & 65.1±0.9          & 78.4±1.9          & 74.8±1.6          & 71.5±1.9          & 68.6±2.9          & 64.1±2.4          \\
    & GCA     & 81.5±0.9          & 76.8±1.0          & 72.0±1.2          & 67.2±1.4          & 61.9±1.8          & 53.4±3.2          & 77.9±1.1          & 75.0±1.3          & 72.7±1.3          & 70.6±1.5          & 67.8±2.3          \\
    & BGRL    & 82.7±1.0          & 77.8±1.2          & 74.8±1.4          & 72.6±1.4          & 69.6±0.8          & 68.0±1.2          & 79.0±0.9          & 76.5±1.3          & 74.2±1.2          & 73.0±0.7          & 70.7±0.8          \\
    & DGI-ADV & 83.7±0.7          & 78.8±1.0          & 76.7±0.7          & 73.8±0.6          & 69.9±1.1          & 68.0±1.4          & 80.6±1.0          & 78.2±1.1          & 75.3±1.8          & 73.2±1.8          & 70.7±2.4          \\
    & ARIEL   & 80.9±0.5          & 75.8±0.8          & 69.8±0.9          & 64.8±1.3          & 60.7±1.5          & 57.6±1.1          & 76.1±1.0          & 70.6±1.1          & 65.4±1.5          & 60.2±1.5          & 53.6±1.7          \\
    \hline
    
    & \proposed 
    & \textbf{83.9±0.7} & \textbf{81.3±1.3} & \textbf{80.2±0.6} & \textbf{78.6±0.4} & \textbf{76.2±1.3} & \textbf{76.8±0.9} & \textbf{81.8±1.3} & \textbf{80.1±1.1} & \textbf{78.7±1.1} & \textbf{77.5±1.5} & \textbf{76.1±1.3} \\
    
    \midrule
    \multirow{5}{*}{Citeseer} 
    & GRACE   & 74.9±0.6          & 72.0±0.7          & 68.8±0.9          & 66.0±0.6          & 63.6±0.8          & 61.3±0.7          & 72.8±0.9          & 71.4±0.7          & 70.1±0.7          & 68.7±0.8          & 67.7±1.1          \\
    & GCA     & 74.2±0.7          & 70.8±0.9          & 67.0±1.6          & 63.6±1.5          & 61.1±1.2          & 57.5±2.2          & 72.3±0.5          & 70.9±0.9          & 69.6±1.1          & 68.5±0.8          & 67.6±0.9          \\
    & BGRL    & 73.4±1.0          & 70.4±1.2          & 67.7±1.0          & 65.0±2.2          & 63.7±1.4          & 61.4±1.7          & 71.5±0.9          & 69.4±0.9          & 68.1±0.7          & 66.6±1.2          & 65.8±1.0          \\
    & DGI-ADV & 76.6±0.3          & 73.1±0.4          & 70.1±0.9          & 67.4±1.0          & 66.0±0.6          & 64.0±0.5          & 74.7±0.5          & 72.8±0.6          & 71.3±0.8          & 69.6±0.4          & 68.2±1.3          \\
    & ARIEL   & \textbf{76.7±0.5} & \textbf{74.2±0.6} & \textbf{72.8±0.8} & 70.2±0.4          & 69.1±0.4          & 67.6±0.7          & \textbf{75.0±0.7} & \textbf{73.7±0.6} & 72.4±0.8          & 71.1±0.8          & \textbf{70.7±0.9} \\
    \hline
    & \proposed  
    & 75.9±0.4          & 74.1±0.7          & 72.7±0.6          & \textbf{70.8±0.8} & \textbf{69.5±0.4} & \textbf{68.3±0.6} & 74.8±0.4          & 73.5±0.6          & \textbf{72.7±0.7} & \textbf{71.7±0.4} & 70.6±0.8          \\
    
    \midrule
    \multirow{5}{*}{Pubmed} 
    & GRACE   & 85.9±0.1          & 82.1±0.2          & 80.1±0.3          & 78.3±0.7          & 76.7±0.3          & 75.7±0.2          & 81.2±0.2          & 78.9±0.1          & 77.3±0.3          & 76.2±0.3          & 75.5±0.2          \\
    & GCA     & \textbf{86.5±0.2} & \textbf{82.6±0.1} & 80.4±0.6          & 78.6±0.7          & 77.1±0.6          & 76.0±0.3          & 81.2±0.2          & 78.6±0.2          & 76.8±0.2          & 75.6±0.3          & 74.8±0.2          \\
    & BGRL    & 85.1±0.2          & 81.3±0.6          & 79.5±0.8          & 78.3±1.0          & 77.2±1.2          & 76.8±0.7          & 80.6±0.8          & 78.7±0.9          & 77.3±1.0          & 76.3±1.2          & 75.6±1.0          \\
    & DGI-ADV & OOM               & OOM               & OOM               & OOM               & OOM               & OOM               & OOM               & OOM               & OOM               & OOM               & OOM               \\
    & ARIEL   & 83.4±0.1          & 79.0±0.4          & 77.2±0.3          & 76.4±0.3          & 75.5±0.2          & 74.8±0.3          & 78.4±0.5          & 76.8±0.3          & 75.7±0.4          & 74.7±0.2          & 74.0±0.5          \\
    \hline              
    & \proposed  
    &  85.5±0.3          & 82.3±0.2          & \textbf{80.7±0.2} & \textbf{79.9±0.1} & \textbf{78.6±0.2} & \textbf{78.0±0.2} & \textbf{82.1±0.2} & \textbf{80.1±0.2} & \textbf{78.7±0.5} & \textbf{77.9±0.4} & \textbf{77.2±0.5} \\
    \bottomrule
\end{tabular}}}
\end{center}
\vspace{-1ex}
\label{tab:main_table3}
\end{table*}

\subsection{Experimental Settings}
\noindent \textbf{Datasets.} \@ We evaluate~\proposed~and baselines on \textbf{thirteen} benchmark datasets, including three citation networks \cite{metattack, nettack}, two co-purchase networks \cite{shchur2018pitfalls}, two co-authorship networks \cite{shchur2018pitfalls}, and six heterophilous networks \cite{geomgcn}. The statistics of the datasets are given in Appendix~\ref{app-sec:dataset}.

\smallskip

\noindent \textbf{Baselines.} \@ The baselines include the state-of-the-art unsupervised graph representation learning (i.e., GRACE, GCA, BGRL) and defense methods (i.e., DGI-ADV, ARIEL). 
Additionally, we include GRACE-MLP, which uses an MLP encoder instead of a GCN encoder, as a baseline that focuses on the node feature information rather than the graph structure.
We describe the details of baseline models in Appendix~\ref{app-sec:baseline}.


\smallskip

\noindent \textbf{Evaluation Protocol.} \@
We evaluate~\proposed~and the baselines under the poisoning attack and evasive attack. The poisoning attack indicates that the graph is attacked before the model training, whereas the evasive attack contains the perturbations only after the model parameters are trained on the clean graph \cite{adv_survey}.
For each setting, we use the public attacked graph datasets offered by \cite{kdd20-prognn} that contain both untargeted attacks \emph{metattack} and targeted attacks \emph{nettack} for the three citation networks. For the co-purchase and co-authorship networks, we create attacked graphs  using \cite{li2020deeprobust} by repeatedly sampling 3,000 nodes and attacking with \emph{metattack} due to the large size of these datasets.

\noindent \textbf{Implementation Details.} \@ The implementation details are described in Appendix~\ref{app-sec:imp_detail}.



\subsection{Performance of Adversarial Defense}

\subsubsection{Node classification under \textit{metattack}}
\label{sec:metattack}
\hfill\\
\noindent \textbf{Setting.} \@ We first train each model in an unsupervised manner, and then evaluate it with the linear evaluation protocol as in \cite{dgi}. We use the same split as in \cite{kdd20-prognn} for the three citation networks, and use the random 1:1:8 split for training, validation, and testing for the co-purchase and the co-authorship networks.

\smallskip

\noindent \textbf{Result.} \@
We first evaluate the robustness of~\proposed~under non-targeted attack (\emph{metattack}). 
In Table~\ref{tab:main_table}, we have the following two observations: 
\textbf{1)} ~\proposed~consistently outperforms other baselines under both the poisoning and the evasive attacks, which indicate that~\proposed~ effectively achieves adversarial robustness with similarity-preserving view and the adversarial view. 
\textbf{2)} The improvements of GRACE-MLP and ~\proposed~ are especially larger under severe perturbation ratios. This implies the benefit of exploiting the node feature information for learning robust representations under severe perturbation ratios.

We further investigate the performance  of~\proposed~and baselines according to the degree of nodes. Fig. \ref{fig:low_degree} shows that \proposed~outperforms all the baselines on low-degree nodes, and particularly outperforms under severe perturbations, which corroborates our theoretical and empirical studies conducted in Section~\ref{subsec:4.2}.
That is, as adversarial attacks on GCL models tend to be generated between low-degree nodes as shown in Section~\ref{subsec:4.2}, it becomes particularly crucial to preserve the node feature similarity for low-degree nodes under severe perturbations. 
In this respect, as~\proposed~focuses on preserving the node feature similarity, it is relatively more robust against attacks on low-degree nodes compared with existing adversarial GCL models.

\begin{figure*}[t!]
  \begin{minipage}{1.25\columnwidth}
  \centering
  \includegraphics[width=1.02\columnwidth]{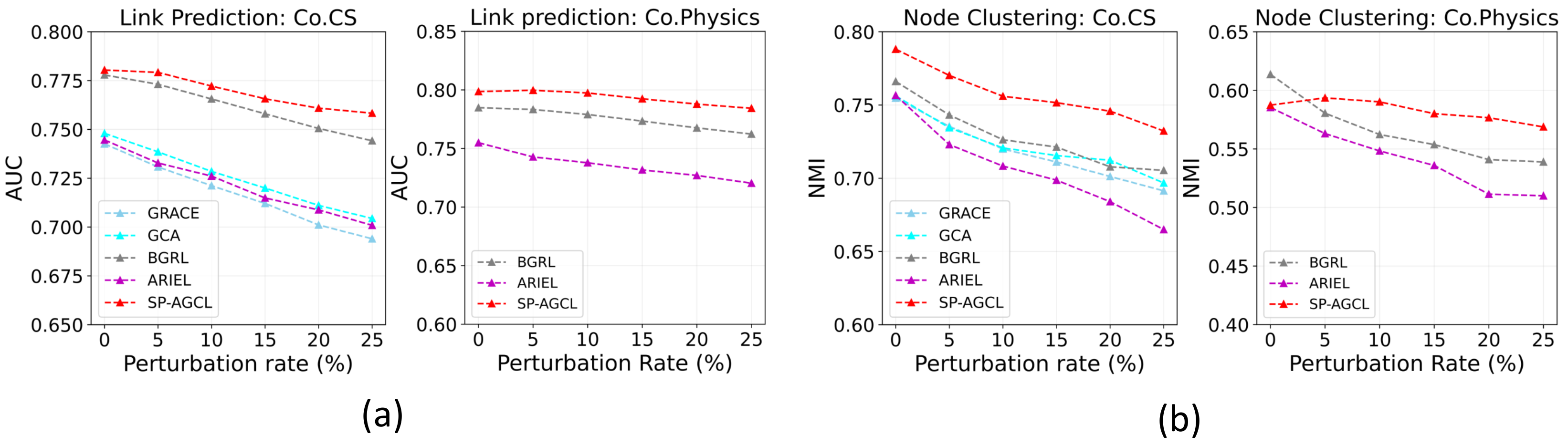}
  \vspace{-5ex}
  \captionof{figure}{(a) Link prediction and (b) Clustering under \emph{metattack} on Co.CS and Co.Physics datasets. (DGI-ADV incurs OOM on both Co.CS and Co.Physics, and GRACE and GCA incurs OOM on Co.Physics dataset.)}
  \label{fig:link_cluster_low_high_degree}
  \end{minipage}
  \vspace{-1.5ex}
  \hfill
  \begin{minipage}{.74\columnwidth}
  \centering
  \includegraphics[width=1.0\columnwidth]{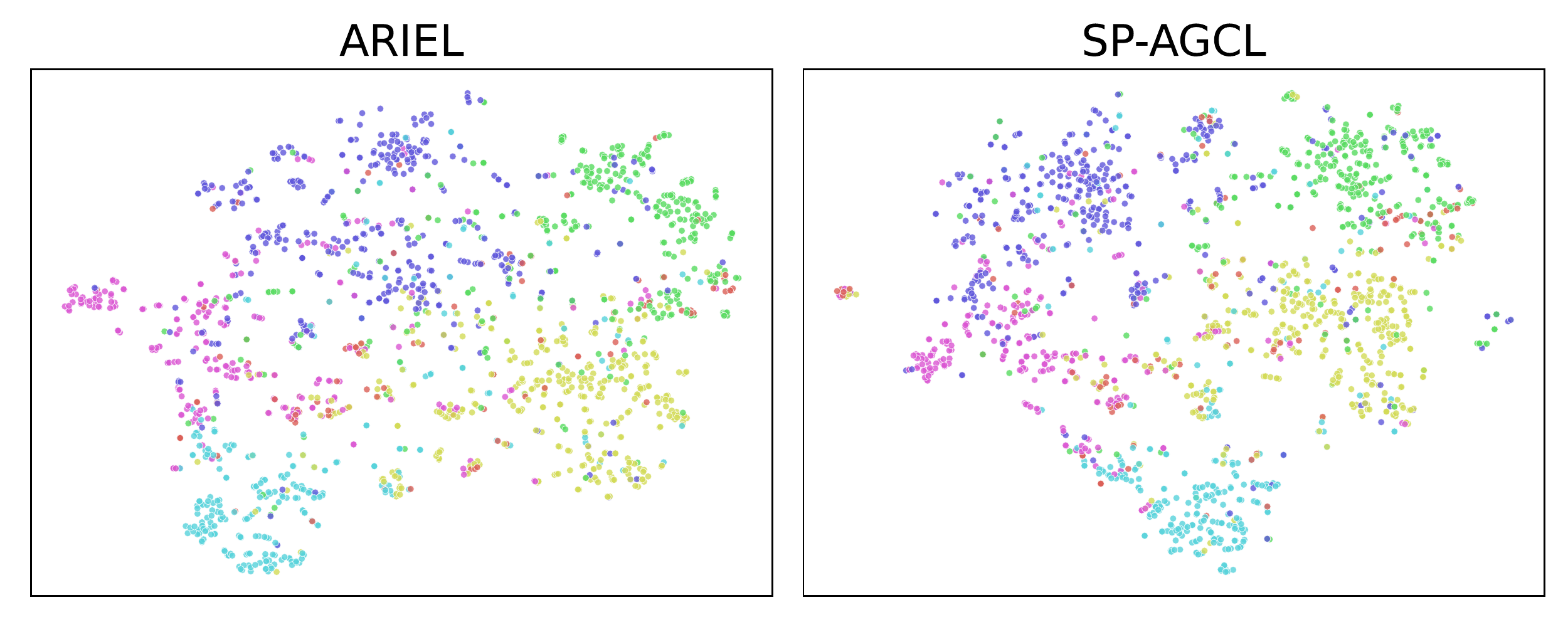}
  \vspace{-4.5ex}
  \caption{t-SNE visualization
  of node representations in Citeseer dataset under \textit{metattack}.}
  \label{fig:tsne}
  \vspace{-3ex}
  \end{minipage}
\end{figure*}

\subsubsection{Node classification under \textit{nettack}} 
\label{sec:nettack}
\hfill\\
\textbf{Setting.} \@ Following the same evaluation procedure described in Section 6.2.1 of the main paper, we train each model (i.e., the node representations) in an unsupervised manner, and then evaluate it with the linear evaluation protocol as in \cite{dgi}. We use the same split and attacked graph structure as in \cite{kdd20-prognn} for the three citation networks, where \emph{nettack} \cite{nettack} is used to generate attacks on specific nodes, which aims at fooling GNNs predictions on those attacked nodes. To be specific, we increase the number of adversarial edges, which are connected to the targeted attacked nodes, from 1 to 5 to consider various targeted attack setups. Then, we evaluate the node classification accuracy on dozens of targeted test nodes with degree larger than 10.

\noindent \textbf{Result.} \@
In Table~\ref{tab:main_table2}, we observe that ~\proposed~ achieves the state-of-the-art performance under targeted attack (\emph{nettack}). Similar to the results in Section. \ref{sec:metattack}, {the performance gap between ~\proposed~ and the baselines gets larger as the number of perturbations for each targeted node increases}. 
The above result implies that exploiting more feature information is helpful for defending the targeted attack, which verifies the effectiveness of ~\proposed.

\subsubsection{Node classification under random perturbations}
\hfill\\
\noindent \textbf{Setting.} \@ We randomly add fake edges into the graph structure to generate attacked graph structure, and then evaluate each model in the same way as above. {We use the three citation networks with the same data split as in \cite{kdd20-prognn} and set the number of added faked edges as from 20\% to 100\% of the number of clean edges.}

\noindent \textbf{Result.} \@
\looseness=-1
In Table~\ref{tab:main_table3}, we observe that~\proposed~consistently outperforms other baselines given randomly perturbed graph structure. Specifically, ~\proposed~demonstrates its robustness under both the poisoning and the evasive attacks setting, showing similar results as reported in Section. \ref{sec:metattack} and Section~\ref{sec:nettack}. The result implies that~\proposed~is robust to random perturbations by preserving the feature similarity and exploiting more feature information.

\subsection{Preserving Feature Similarity}
\noindent \textbf{Analysis on Feature Similarity.} \@ To verify whether~\proposed~preserves the node feature similarity, we compute $OL$ scores of~\proposed~in Fig. \ref{fig:observation}(b) (Refer to Eqn.~(\ref{eq:ol})). We observe that~\proposed~has the highest $OL$ scores on Citeseer and Co.CS datasets. This implies that~\proposed~preserves the feature similarity information unlike the adversarially trained GCL model whose node representations lose the feature similarity.

\smallskip
\noindent \textbf{Benefit of Preserving Feature similarity.} \@
We further investigate the benefits of preserving feature similarity in other downstream tasks (i.e., link prediction \& node clustering). Fig.~\ref{fig:link_cluster_low_high_degree}(a) and (b) show the result on Co.CS and Co.Physics datasets under \emph{metattack}.
For \textit{link prediction}, we closely follow a commonly used setting~\cite{zhang2018link,asp2vec} and use area under curve (AUC) as the evaluation metric. 
We observe that \proposed~consistently predicts reliable links compared with other baselines across all the perturbation ratios. Furthermore, ARIEL, the state-of-the-art adversarial GCL model, shows the worst performance. We argue that node feature information is beneficial to predicting reliable links since nodes with similar features tend to be adjacent in many real-world graphs \cite{rsgnn, kdd20-prognn}. Hence, our proposed adversarial feature masking and similarity-preserving view play an important role in predicting reliable links since they make the node representations retain more feature information. 
For \textit{node clustering}, we perform $k$-means clustering on the learned node representations, where $k$ is set to the number of classes, to verify whether the clusters are separable in terms of the class labels. We use normalized mutual information (NMI) as the evaluation metric.
We observe that~\proposed~consistently outperforms ARIEL in node clustering as well, which demonstrates that preserving the node feature information is crucial as it is highly related to class information.

\smallskip
\noindent \textbf{Visualization of Representation Space.} \@ We visualize the node representations of ARIEL and \proposed~ via t-SNE \cite{tsne} to intuitively see the effect of preserving the node feature information in the representation space. Fig. \ref{fig:tsne} shows the node representations of ARIEL and \proposed~ trained on a citeseer graph under 25\% \textit{metattack}. We observe that the representations of ARIEL are separable but widely distributed resulting in vague class boundaries. On the other hand, the representations of \proposed~ are more tightly grouped together, resulting in more separable class boundaries. We attribute such a difference to the AT of ARIEL that incurs a loss of node feature information, which is preserved in \proposed. Furthermore, the vague class boundaries of ARIEL explain the poor performance of ARIEL in the node clustering task shown in Fig. \ref{fig:link_cluster_low_high_degree}(b).


\subsection{Experiments on Real-World Scenarios}
\begin{figure}[t]
    \centering
    \includegraphics[width=0.65\linewidth]{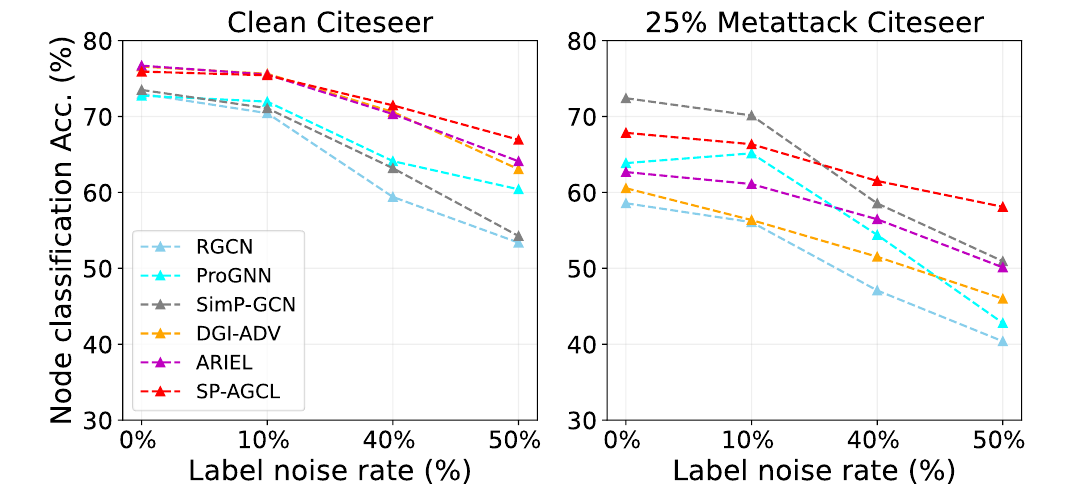}
    \vspace{-2ex}
    \caption{Node classification accuracy with noisy label.}
    \label{fig:noisy}
    \vspace{-4ex}
\end{figure}

\noindent \textbf{Node classification with Noisy Label.} \@ 
We compare ~\proposed~ with both supervised (i.e., RGCN~\cite{kdd-19-rgcn}, ProGNN~\cite{kdd20-prognn}, and SimP-GCN~\cite{simpgcn}) and unsupervised defense methods (i.e., DGI-ADV and ARIEL) to confirm the effectiveness of models when the label information contains noise. We train the models on the clean and poisoned Citeseer dataset, whose label noise rates are varied. In Fig. \ref{fig:noisy}, we observe that the unsupervised methods (especially~\proposed) outperform the supervised methods at relatively high noise rates (i.e., 40\% and 50\%). This is because the node representations of the supervised defense methods are not well generated in terms of the downstream task since they heavily rely on the supervision signals obtained from the noisy node labels. 
Moreover, we observe that~\proposed~outperforms unsupervised methods (i.e., DGI-ADV and ARIEL), and the performance gap gets larger as the label noise rate increases. 
We argue that this is mainly because~\proposed~better exploits feature information, which results in more robust node representations under noisy labels demonstrating practicality of~\proposed~in reality.

\begin{table}[h]
    \centering
    \vspace{-1ex}
    \caption{Node classification Acc. on heterophilous graphs.}
    \vspace{-2ex}
    \centering
    \resizebox{0.88\linewidth}{!}{
    \begin{tabular}[b]{l|cccccc}
    \hline 
    & Chameleon & Squirrel & Actor & Texas & Wisconsin & Cornell\\
    \hline
    GRACE   & 46.6±2.8          & 35.2±1.0            & 29.5±0.5          & 61.1±6.5          & 55.3±5.5 & 61.1±5.0          \\
GCA     & 50.0±3.0              & 37.1±1.8          & 29.3±0.8          & 60.0±6.3            & 55.7±8.0   & 59.5±3.8          \\
BGRL    & 57.1±3.6          & 40.6±1.6          & 31.0±1.2            & 61.6±6.0            & 57.7±5.2 & 57.8±4.7          \\
DGI-ADV & 53.4±2.2          & 40.1±1.6          & 26.5±0.9          & 58.4±6.1          & 57.3±4.9 & 60.5±5.8          \\
ARIEL   & 44.3±2.4          & 36.8±1.2          & 29.6±0.3          & 58.4±4.7          & 53.3±7.2 & 57.8±4.4          \\
    \hline
    \proposed         & \textbf{57.5±2.5} & \textbf{41.1±1.9} & \textbf{32.3±1.3} & \textbf{64.9±6.8} & \textbf{58.4±5.5}      & \textbf{64.3±3.6} \\ 
    \hline
    \end{tabular}}
    \vspace{-1ex}
    \label{tab:hetero}
\end{table}

\smallskip
\noindent \textbf{Node classification on Heterophilous Networks.} \@ 
Throughout this paper, we showed that preserving node feature similarity is crucial when graphs are poisoned/attacked. In fact, a heterophilous network~\cite{zhu2021graph} in which nodes with dissimilar properties (e.g., node features and labels) are connected can be considered as a poisoned/attacked graph considering the behavior of adversarial attacks described in Section~\ref{sec:4}. Hence, in this section, we evaluate~\proposed~on six commonly used heterophilous networks benchmark in terms of node classification. In Table~\ref{tab:hetero}, we observe that~\proposed~outperforms baselines on heterophilous networks.
Moreover, ARIEL, which is an adversarially trained variant of GRACE, performs worse than GRACE. This implies that adversarial training fails to preserve the feature similarity, and that the feature similarity should be preserved when the given structural information is not helpful (as in heterophilous networks).

\smallskip

\subsection{Ablation Study}
\label{sec:ablation}
To evaluate the importance of each component of~\proposed, i.e., similarity-preserving view (\textit{SP}), and the feature perturbation in adversarial view generation (\textit{Feat. Ptb}), we incrementally add them to a baseline model, which is GRACE~\cite{grace} with structural perturbations described in Section \ref{sec:adv_view}.1.
As for the adversarial view generation, we compare our proposed feature masking strategy with the feature flipping strategy~\cite{grv,ariel}.
We have the following observations in Table \ref{tab:ablation}.
\textbf{1) } Adding the similarity-preserving view is helpful, especially under severe structural perturbations, which demonstrates the benefit of preserving the node feature similarity in achieving adversarial robustness against graph structural attacks.
\textbf{2) } When considering the adversarial view, adding the feature masking component is helpful in general, which again demonstrates the importance of exploiting the node feature information.
\textbf{3) } Comparing the strategies for the feature perturbations, our proposed masking strategy outperforms the flipping strategy, even though the masking strategy requires less computations. We attribute this to the fact that the feature flip greatly alters the co-occurrence/correlation statistics of node features, which incurs an adverse effect on the AT by making the clean view and the adversarial view too distant from each other.

\begin{table}
    \centering
    \caption{Ablation study. \textit{SP} and \textit{Feat. Ptb} denote the existence of similarity-preserving view and the type of feature perturbations, respectively.}
    \vspace{-2ex}
    \scalebox{0.78}{
    \renewcommand{\arraystretch}{0.9}
    \begin{tabular}{cc|ccc|ccc}
        \hline
        \multicolumn{2}{c|}{Component} & \multicolumn{3}{c}{Cora} & \multicolumn{3}{c}{Citeseer}  \\
        \hline
        SP & Feat. Ptb & 0\% & 15\% & 25\% & 0\% & 15\% & 25\% \\
        \hline
         \ding{55} & \ding{55} & 82.9±1.1 & 64.8±1.0 & 50.6±1.0 & 70.6±1.1 & 63.3±1.4 & 55.3±2.4\\
         \ding{55} & Flip  & 83.4±0.7 & 60.5±1.5 & 48.0±1.3 & 72.5±1.1 & 63.8±1.3 & 52.6±1.9 \\
         \ding{55} & Mask  & \textbf{84.0±1.0} & \textbf{68.8±1.3} & \textbf{52.0±0.7} & \textbf{74.4±1.0} & \textbf{70.0±1.6} &	\textbf{64.1±0.8}\\
         \midrule
         \midrule
         \ding{51} & \ding{55} & 83.8±0.9 & 68.4±1.2 & 64.7±1.4 & 74.5±0.5 & 69.7±0.9 & 68.6±0.9\\
        \ding{51} & Flip & 82.0±0.5 & 67.5±1.1 &  62.7±2.7 & 73.7±0.5 & 68.0±1.3 & 62.8±0.9\\
        \ding{51} & Mask & \textbf{83.9±0.7} & \textbf{73.3±0.5} &  \textbf{65.3±1.0} & \textbf{75.9±0.4} & \textbf{72.1±1.1} & \textbf{69.6±0.9}\\
        \hline
    \end{tabular}}
    \label{tab:ablation}
\vspace{-2ex}
\end{table}


\subsection{Complexity Analyses}


\noindent \textbf{Analysis on Computational Efficiency} \@
We compare the training time and the attacked view generation time of ARIEL, and \proposed~ on Co.Physics dataset. For a fair comparison, ARIEL and \proposed~utilize the same size of subgraphs during training. In Table~\ref{tab:converge} and Fig.~\ref{fig:converge}, we observe that \proposed~is faster than ARIEL in terms of both the training and the view generation. This implies that our proposed adversarial view generation is scalable compared with the PGD attack \cite{ijacai-19-pgd-topology-attack-defense} used in ARIEL, which creates an adversarial view through repeated iterations and complex optimization. In addition to the fast model training, Fig. \ref{fig:converge} shows that \proposed~ converges faster than ARIEL. As a result, \proposed~is proven to be the most efficient and effective method.

\begin{minipage}{.6\columnwidth}
\centering
\small
\captionof{table}{Averaged running time of 100 duration  over 4000 epochs on Co.Physics dataset.}
\label{tab:converge}
\scalebox{0.87}{
\renewcommand{\arraystretch}{0.9}
\begin{tabular}{c|cc}
\noalign{\smallskip}\noalign{\smallskip}\hline
 &  ARIEL & \proposed  \\
\hline
Time (s) / 100 attacks & 7.5 & \textbf{1.8} \\
Time (s) / 100 epochs & 12  & \textbf{7.2} \\
\hline
\end{tabular}
}
\end{minipage}
\hfill%
\begin{minipage}{.38\linewidth}
\vspace{-2ex}
\begin{figure}[H]
\centering
\includegraphics[width=.80\linewidth]{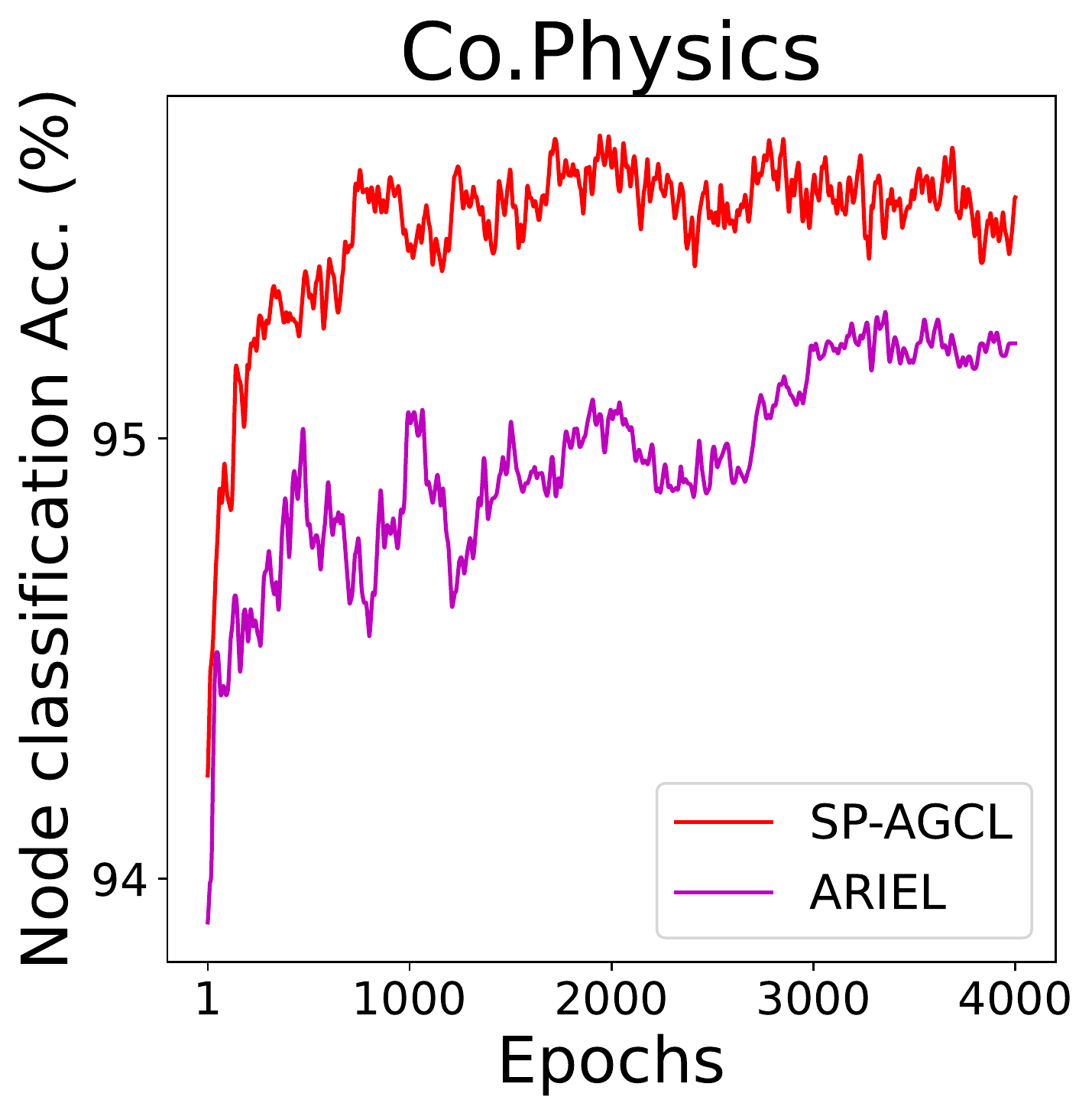}
\end{figure}
\vspace{-5ex}  
\captionof{figure}{Valid. Acc. of node classification on Co.Physics dataset.}
\vspace{-2ex}  
\label{fig:converge}

\end{minipage}

\section{Conclusions}
In this paper, we discover that adversarial GCL models obtain robustness against adversarial attacks at the expense of not being able to preserve the node feature similarity information through theoretical and empirical studies. Based on our findings, we propose~\proposed~that learns robust node representations that preserve the node feature similarity by introducing the similarity-preserving view. Moreover, the proposed adversarial feature masking exploits more feature information.
We verify the effectiveness of~\proposed~by conducting extensive experiments on thirteen benchmark datasets with multiple attacking scenarios along with several real-world scenarios such as networks with noisy labels and heterophily.

\section*{Acknowledgements}

This work was supported by Institute of Information \& communications Technology Planning \& Evaluation (IITP) grant funded by the Korea government(MSIT) (No.2022-0-00157 and No.2022-0-00077).

\bibliographystyle{ACM-Reference-Format}
\balance
\bibliography{refer.bib}

\appendix

\clearpage

\section{Overall Framework of~\proposed}
\label{app-sec:overall_framework}

\begin{figure}[h]
    \includegraphics[width=.8\columnwidth]{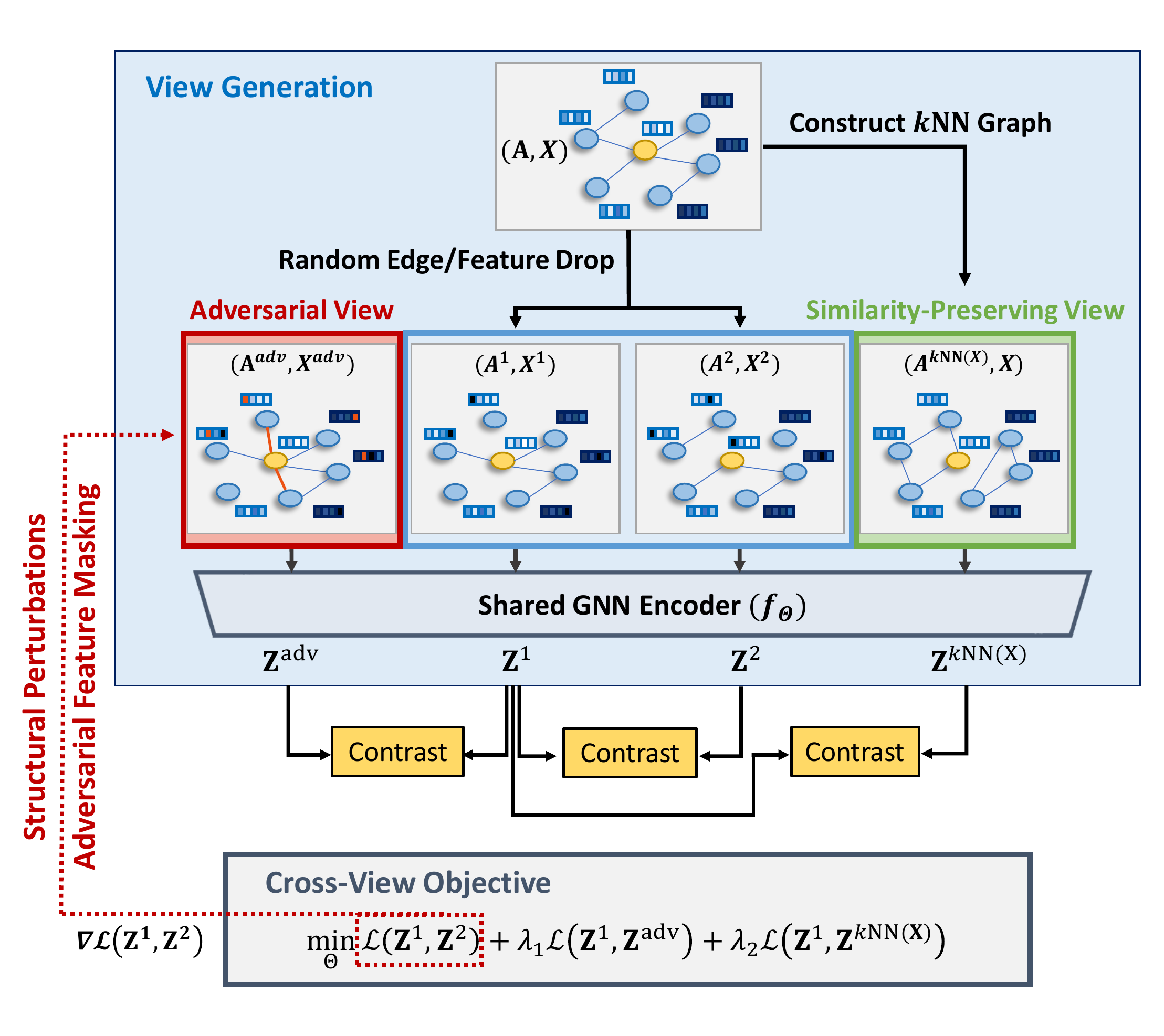}
    \vspace{-4ex}
    \caption{The overall architecture of~\proposed.}
    \label{app-fig:architecture}
    \vspace{-2ex}
\end{figure}

\section{Datasets.}
\label{app-sec:dataset}
We evaluate~\proposed~and baselines on \textbf{thirteen} benchmark datasets, including three citation networks \cite{metattack, nettack}, two co-purchase networks \cite{shchur2018pitfalls}, two co-authorship networks \cite{shchur2018pitfalls}, and six heterophilous networks \cite{geomgcn}. The statistics of the datasets are shown in Table. \ref{tab:dataset}.

\begin{table}[h]
\centering
\small
\caption{Statistics for datasets.}
\vspace{-2ex}
\renewcommand{\arraystretch}{0.98}
{\small
\scalebox{0.9}{
\begin{tabular}{c|c|cccc}
\hline
 Domain & Dataset & \# Nodes & \# Edges & \# Features & \# Classes \\
\hline 
\multirow{3}{*}{Citation} & Cora     & 2,485          & 5,069         & 1,433      & 7   \\  
& Citeseer   & 2,110        & 3,668         & 3,703      & 6   \\
& Pubmed     & 19,717       & 44,338        & 500        & 3   \\
\hline 
\multirow{2}{*}{Co-purchase} & Am.Photo   & 7,650        & 119,081       & 745        & 8   \\
& Am.Comp    & 13,752       & 245,861       & 767        & 10  \\
\hline 
\multirow{2}{*}{Co-author}& Co.CS      & 18,333       & 81,894        & 6,805      & 15  \\
& Co.Physics   & 34,493     & 247,962       & 8,415      & 5    \\
 \hline

\multirow{6}{*}{Heterohpily} & Chameleon   &    2,277      &   36,101        &    2,325 & 5    \\
& Squirrel   &    5,201      &   217,073        &    2,089       &       5    \\
& Actor   &    7,600      &   33,544        &    931       &       5    \\
& Cornell   &    183      &   295        &    1,703       &       5    \\
& Texas   &    183      &   309        &    1,703       &       5    \\
& Wisconsin   &    251      &   499        &    1,703       &       5    \\
\hline

\end{tabular}
}}
\label{tab:dataset}
\vspace{0ex}
\end{table}

\section{Details on Experimental Settings.}

\subsection{Baselines.}
\label{app-sec:baseline}

The baselines include the state-of-the-art unsupervised graph representation learning methods (i.e., GRACE, GCA, BGRL) and defense methods (i.e., DGI-ADV, ARIEL) that adopt adversarial training (AT). 

\begin{itemize}
    \item \noindent\textbf{GRACE} \cite{grace} \@ is an augmentation-based GCL model that pulls the representations of two semantically similar nodes and pushes that of two semantically different nodes.

    \item \noindent\textbf{GCA} \cite{gca} \@ is a more advanced method of GRACE that employs various augmentation strategies.

    \item \noindent\textbf{BGRL} \cite{bgrl} is the state-of-the-art graph representation learning method that minimizes the distance of two node representations from different augmentations without negative samples.

    \item \noindent\textbf{DGI-ADV} \cite{grv} is trained by alternatively optimizing DGI \cite{dgi} and AT objective based on the vulnerability of the graph representation.
    
    \item \noindent\textbf{ARIEL} \cite{ariel} is the state-of-the-art adversarial GCL model that adopts AT to GRACE framework with an information regularizer for stabilizing the model training.

\end{itemize}

\subsection{Implementation Details}
\label{app-sec:imp_detail}
For each experiment, we report the average performance of 10 runs with standard deviations. For GRACE, GCA, BGRL, and ARIEL, we use the best hyperparameter settings presented in their papers. For DGI-ADV, we use the default hyperparameter settings in its implementation.

For \proposed, we tune the learning rate and weight decay from \{0.05, 0.01, 0.005, 0.001\} and \{0.01, 0.001, 0.0001, 0.00001\}, respectively. For generating two augmented views, we tune drop edge/feature rates from \{0.1, 0.2, 0.3, 0.4, 0.5\}. For generating the adversarial view, we tune edge perturbation budget $\Delta_{\mathbf{A}}$ and feature masking ratio  $\Delta_{\mathbf{X}}$ from \{0.1, 0.3, 0.5, 0.7, 0.9\} and \{0.0, 0.1, 0.3, 0.5, 0.7, 0.9\}, respectively. For generating the similarity-preserving view, we tune $k$ of $k$NN graph from \{5, 10\}. Moreover, we tune the combination coefficient $\lambda_1$ and $\lambda_2$ from \{0.1, 0.5, 1,2,3,4,5\}.

\section{Additional Experimental Results.}

In this section, we conduct three additional experiments to analyze the sensitivity of~\proposed~to the hyperparameters including 1) the perturbation budgets of the graph structure and node features $\Delta_{\mathbf{A}}$, $\Delta_{\mathbf{X}}$, 2) the coefficients of objective $\lambda_1$, $\lambda_2$, and 3) $k$ value of the $k$NN algorithm in similarity-preserving view generation.

\vspace{-2ex}
\subsection{Sensitivity analysis on $\Delta_{\mathbf{A}}$ and $\Delta_{\mathbf{X}}$}
\label{sec:b3}
\begin{figure*}[h]
    \centering
    \vspace{1ex}  
    \includegraphics[width=0.89\linewidth]{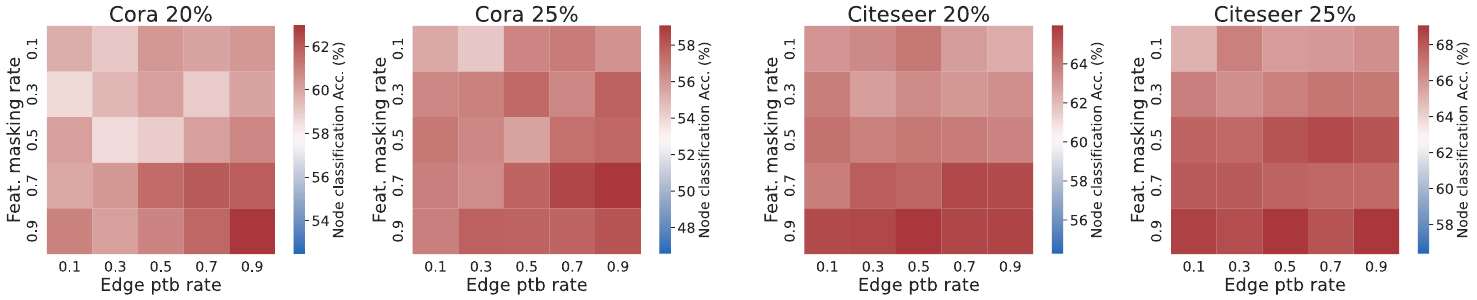}
    \vspace{-2ex}
    \caption{Sensitivity analysis on edge perturbation and feature masking rates. \textcolor{burntumber}{Red}-white-\textcolor{airforceblue}{blue} means outperformance, on-par, and underperformance compared with ARIEL.}
    \label{fig:sensitivity_ptb_bdgt}
    \vspace{-3ex}
\end{figure*}

\looseness=-1
We analyze the sensitivity of the perturbation budgets for generating the adversarial view. Specifically, we change the edge perturbation ratio $\Delta_{\mathbf{A}}$ and the feature masking ratio $\Delta_{\mathbf{X}}$ from 0.1 to 0.9 to confirm that the performance of~\proposed~is insensitive to the perturbation ratio as AT is performed. In Figure \ref{fig:sensitivity_ptb_bdgt}, we observe that the performance of~\proposed~is consistently better than that of ARIEL in terms of the accuracy on attacked Cora and Citeseer datasets. This implies that~\proposed~is not sensitive to the perturbation budgets, showing that both perturbing the graph structure and masking the feature are helpful for enhancing the robustness of~\proposed.

\subsection{Sensitivity analysis on $\lambda_1$ and $\lambda_2$.}

\begin{figure*}[t]
    \centering
    \vspace{1ex}  
    \includegraphics[width=0.88\linewidth]{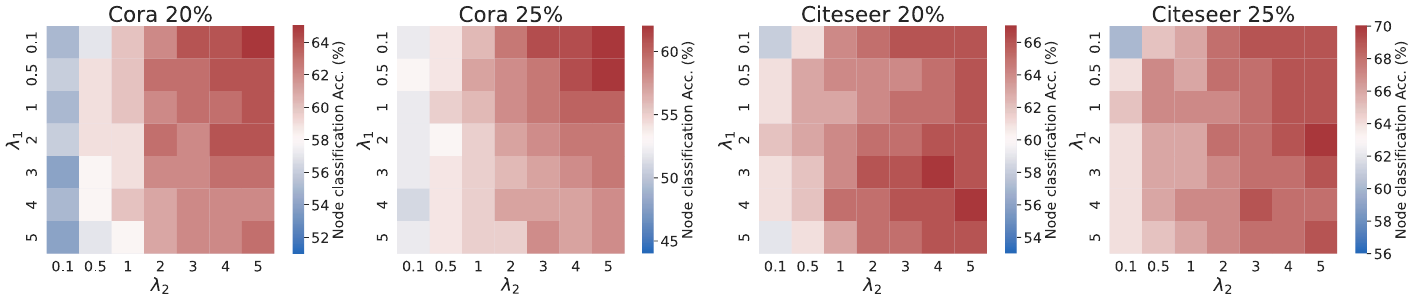}
    \vspace{-2ex}
    \caption{Sensitivity analysis on the coefficients of objectives $\lambda_1$ and $\lambda_2$. \textcolor{burntumber}{Red}-white-\textcolor{airforceblue}{blue} means outperformance, on-par, and underperformance compared with ARIEL.}
    \label{fig:sensitivity_labmda}
    \vspace{-3ex}
\end{figure*}

We analyze the sensitivity of the coefficients of the training objective $\lambda_1$ and $\lambda_2$ in Eqn (6) of the main paper. Note that $\lambda_1$ determines the importance of the adversarial view, while $\lambda_2$ determines the importance of the similarity-preserving view.  We conduct a grid search for the two hyperparameters with the values in $\{0.1, 0.5, 1, 2, 3, 4, 5\}$. In Figure \ref{fig:sensitivity_labmda}, we observe that \proposed~generally outperforms ARIEL in terms of the accuracy on attacked Cora and Citeseer datasets, showing that ~\proposed~is not senstivie to the selection of $\lambda_1$ and $\lambda_2$. However, when $\lambda_2$ is relatively small (i.e., $\lambda_2=0.1,0.5$), ~\proposed~shows comparable or worse performance than ARIEL. In other words, ~\proposed~becomes more robust when the similarity-preserving view plays more significant role in the learning objective (i.e., when $\lambda_2$ is large). This implies that the similarity-preserving view is important for achieving robustness of~\proposed.

\begin{figure*}[t]
    \centering
    \vspace{1ex}  
    \includegraphics[width=0.88\linewidth]{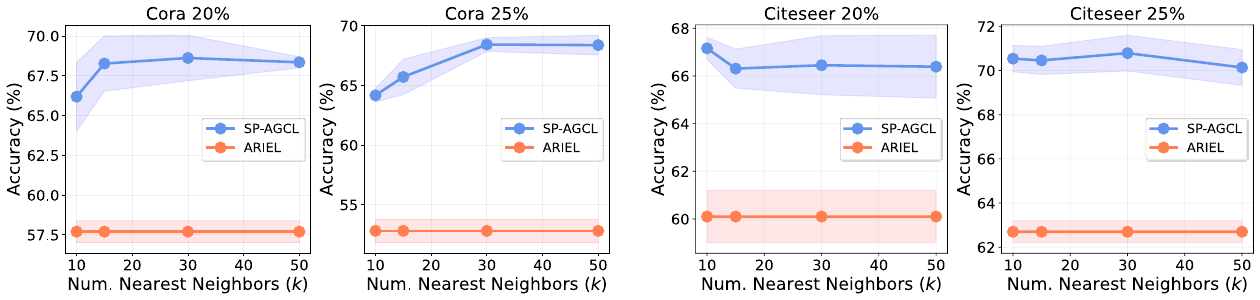}
    \vspace{-2ex}
    \caption{Sensitivity analysis on $k$NN.}
    \label{fig:sensitivity_k}
    \vspace{-3ex}
\end{figure*}

\subsection{Sensitivity analysis on $k$NN.}
We analyze the sensitivity of~\proposed~over the number of nearest neighbors (i.e., $k$ value of $k$NN) for generating the similarity-preserving view. To be specific, we increase the value of $k$ from 10 to 50 (i.e., \{10,15,30,50\}), and evaluate the accuracy of~\proposed~and ARIEL. In Figure \ref{fig:sensitivity_k}, we observe that the performance of~\proposed~is not only insensitive to $k$ value, but also greatly outperforms ARIEL regardless of the value of $k$ in terms of the accuracy on attacked Cora and Citeseer datasets. Moreover, this implies that employing the node feature similarity is helpful for learning robust representations of GCL models regardless of the $k$ values.

\section{Detailed Proof of Equation \ref{eq:diff}.}
\label{eqn:proof}
We provide the detailed derivations of Eqn. (\ref{eq:diff}). Note that the representation $\mathbf{z}_i$ after 1 layer GCN is applied can be computed as follow: $\mathbf{z}_i=\sum_{j\in\mathcal{N}_{\mathbf{A}}^i\cup\ \{i\}}\frac{\mathbf{W}x_j}{\sqrt{|\mathcal{N}_{\mathbf{A}}^i|}{\sqrt{|\mathcal{N}_{\mathbf{A}}^j|}}}$ where $\mathbf{x}_j$ is the input feature for node $j$, $\mathcal{N}_{\mathbf{A}}^i$ is the neighbor set of node $i$ give the adjacency matrix $\mathbf{A}$, and $|\cdot|$ counts the number of element in a given set. Then, the difference between the clean representation and attacked representation is given as follows:
{\small
\begin{align*}
    & \mathbf{z}_i^2 - {\mathbf{z}_i^{\text{atk}}} = (\mathbf{z}_i^2 - \mathbf{z}_i^1) + (\mathbf{z}_i^1 -{\mathbf{z}_i^{\text{atk}}}) \nonumber \\ 
    &= \mathbf{e}_i \!+ \!\!\!\!\!\sum_{j\in\mathcal{N}_{\mathbf{A}^1}^i\cup \{i\}} \frac{{\mathbf{W}\mathbf{x}_j}}{\sqrt{|\mathcal{N}_{\mathbf{A}^1}^i|}\sqrt{|\mathcal{N}_{\mathbf{A}^1}^j|}} - \!\!\!\!\!\sum_{j\in  \mathcal{N}_{\mathbf{A}^1+\delta_{\mathbf{A}}}^i \!\!\!\cup \{i\}} \frac{{\mathbf{W}\mathbf{x}_j}}{\sqrt{|\mathcal{N}_{\mathbf{A}^1+\delta_{\mathbf{A}}}^i|}\sqrt{|\mathcal{N}_{\mathbf{A}^1+\delta_{A}}^j|}} \nonumber \\
    &= \mathbf{e}_i \!+ \!\!\!\!\!\sum_{j\in\mathcal{N}_{\mathbf{A}^1}^i\cup \{i\}} \frac{{\mathbf{W}\mathbf{x}_j}}{\sqrt{|\mathcal{N}_{\mathbf{A}^1}^i|}\sqrt{|\mathcal{N}_{\mathbf{A}^1}^j|}} \\
    & - \left(\sum_{j\in  \mathcal{N}_{\mathbf{A}^1}^i \!\!\!\cup \{i\}} \frac{{\mathbf{W}\mathbf{x}_j}}{\sqrt{|\mathcal{N}_{\mathbf{A}^1}^i|+1}\sqrt{|\mathcal{N}_{\mathbf{A}^1}^j|}} + \frac{\mathbf{W}\mathbf{x}_k}{\sqrt{|\mathcal{N}_{\mathbf{A}^1}^i|+1}\sqrt{|\mathcal{N}_{\mathbf{A}^1}^k|+1}}\right) \nonumber \\
    &= \mathbf{e}_i +  \sum_{j\in\mathcal{N}_{\mathbf{A}^1}^i\cup \{i\}} \frac{\mathbf{W}\mathbf{x}_j}{\sqrt{|\mathcal{N}_{\mathbf{A}^1}^j|}} \left( \frac{\sqrt{|\mathcal{N}_{\mathbf{A}^1}^i|+1}-\sqrt{|\mathcal{N}_{\mathbf{A}^1}^i|}}{\sqrt{|\mathcal{N}_{\mathbf{A}^1}^i|}\sqrt{|\mathcal{N}_{\mathbf{A}^1}^i|+1}}\right) 
    - \frac{\mathbf{W}\mathbf{x}_k}{\sqrt{|\mathcal{N}_{\mathbf{A}^1}^i|+1}\sqrt{|\mathcal{N}_{\mathbf{A}^1}^k|+1}}\\
    &= \mathbf{e}_i + \frac{1}{\underbrace{\sqrt{|\mathcal{N}_{\mathbf{A}^1}^i|+1}}_{\text{Degree term}}} \underbrace{\left(\sum_{j\in\mathcal{N}_{\mathbf{A}^1}^i\cup \{i\}} \frac{\alpha{\mathbf{W}\mathbf{x}_j}}{\sqrt{|\mathcal{N}_{\mathbf{A}^1}^i|}\sqrt{|\mathcal{N}_{\mathbf{A}^1}^j|}} - \frac{\mathbf{W}{\mathbf{x}_k}}{\sqrt{|\mathcal{N}_{\mathbf{A}^1}^k|+1}}\right)}_{\text{Feature difference term}}
    \label{eq:diff}
\end{align*}
}%

\citestyle{acmauthoryear}

\end{document}